\pdfoutput=1 
\documentclass[11pt]{article}

\usepackage{acl}
\hypersetup{hidelinks}
\usepackage{times}
\usepackage{latexsym}
\usepackage[T1]{fontenc}
\usepackage[utf8]{inputenc}
\usepackage{microtype}
\usepackage{booktabs}
\usepackage{graphicx}
\usepackage{mdframed}
\usepackage{amsmath}
\usepackage[capitalize]{cleveref} 
\usepackage{newtxmath} 
\usepackage{caption,subcaption} 

\newcommand{\bpe}{\scalebox{0.8}{@@}\ }

\newcommand{\eos}{\texttt{EOS}}

\newcommand{\ucg}{UCG}

\title{A Continuum of Generation Tasks for Investigating\\Length Bias and Degenerate Repetition}

\author{Darcey Riley \and David Chiang \\
        University of Notre Dame \\
        \texttt{\{darcey.riley,dchiang\}@nd.edu}}

\let\cite\undefined

\begin{document}
\maketitle
\begin{abstract}
Language models suffer from various degenerate behaviors. These differ between tasks: machine translation (MT) exhibits length bias, while tasks like story generation exhibit excessive repetition. Recent work has attributed the difference to \emph{task constrainedness}, but evidence for this claim has always involved many confounding variables.
To study this question directly, we introduce a new experimental framework that allows us to smoothly vary task constrainedness, from MT at one end to fully open-ended generation at the other, while keeping all other aspects fixed. We find that: 
(1) repetition decreases smoothly with constrainedness, explaining the difference in repetition across tasks;
(2) length bias surprisingly also decreases with constrainedness, suggesting some other cause for the difference in length bias;
(3) across the board, these problems affect the mode, not the whole distribution; (4) the differences cannot be attributed to a change in the entropy of the distribution, since another method of changing the entropy, label smoothing, does not produce the same effect.
\end{abstract}

\section{Introduction}

Neural language models serve as the core of modern NLP technologies, but they suffer from ``inadequacy of the mode'' \citep{eikema-aziz-2020-map,zhang-etal-2021-trading}, in which the sentences with the very highest probability under the model exhibit various pathological behaviors. Specifically, machine translation suffers from \emph{length bias}, where the generated translations are too long or (more often) too short \citep{murray-chiang-2018-correcting,stahlberg-byrne-2019-nmt}, while story generation suffers from \emph{degenerate repetition}, where the generated text repeats words or phrases unnecessarily \citep{holtzman-etal-2020-curious}.

It has frequently been assumed that length bias and degenerate repetition are both aspects of a single phenomenon;
for instance, it is very common for papers studying MT to reference issues observed in story generation. However, MT and story generation exhibit very different problems, and have been addressed using very different solutions.
So it is worth pausing for a moment to ask how they relate. Are they truly two symptoms of the same problem? Why do different tasks exhibit different degenerate behaviors?

\citet{stahlberg-etal-2022-uncertainty} and \citet{wiher-etal-2022-decoding} 
attribute the differences to \textit{task constrainedness}: given a particular input, how many different possible correct answers might there be? For example, grammatical error correction (GEC) and speech recognition are more constrained; image captioning and MT are in the middle; and story generation, dialogue, and pure unconditioned generation from the language model (\ucg{}) are least constrained.
However, constrainedness is only one of many differences among these tasks. They also differ in the length of the inputs and outputs, the size of the models, and so on. So although these papers provide compelling circumstantial evidence that the differences can be explained in terms of constrainedness, they do not rule out alternative hypotheses.

In this paper, we introduce a new experimental framework which lets us directly adjust constrainedness  while keeping everything else (architecture, number of parameters, type of output data) fixed. We expect that, if task constrainedness really is responsible for the differences in degenerate behaviors seen across tasks, 
then these behaviors should vary smoothly as we adjust constrainedness.
We find this to be true for degenerate repetition: we see basically none for pure MT, and an increasing amount as we lower the constrainedness down to \ucg{}. This is consistent with the literature, which reports repetition as a problem in \ucg{}, but not~MT\@.

On the other hand, for length bias, we discover, to our knowledge for the first time, that length bias actually \textit{increases} for less constrained tasks. This is inconsistent with the literature, where length bias is commonly reported for MT but very rarely reported for \ucg{}. We conclude that the difference, then, is either due to some other factor besides constrainedness influencing the model's probability distribution, or that it can be attributed to the different decoding strategies commonly used for the different tasks.

In addition, we present results showing that both length bias and degenerate repetition are problems exclusive to the mode; they do not in general affect random samples from the distribution. Lastly, we explore one possible explanation for why length bias and repetition differ across constrainedness levels: that it is because less constrained tasks have higher entropy. We find that this cannot be the explanation, as another method of increasing the entropy, label smoothing, has very little effect on these phenomena.

\begin{table*} \centering \scriptsize
\begin{tabular}{@{}rrl@{}}
\toprule
$s$ (\%) & length & tokens \\
\midrule
0 & 0 & \eos{} \\
10 & 3 & Sch\bpe{}on heute \eos{} \\
20 & 7 & Sch\bpe{}on heute spru\bpe{}d\bpe{}elt in \eos{} \\
30 & 8 & Sch\bpe{}on heute spru\bpe{}d\bpe{}elt in einigen \eos{} \\
40 & 13 & Sch\bpe{}on heute spru\bpe{}d\bpe{}elt in einigen f\bpe{}la\bpe{}chen Se\bpe{}en \eos{} \\
50 & 14 & Sch\bpe{}on heute spru\bpe{}d\bpe{}elt in einigen f\bpe{}la\bpe{}chen Se\bpe{}en in \eos{} \\
60 & 19 & Sch\bpe{}on heute spru\bpe{}d\bpe{}elt in einigen f\bpe{}la\bpe{}chen Se\bpe{}en in Al\bpe{}as\bpe{}ka Me\bpe{}than \eos{} \\
70 & 21 & Sch\bpe{}on heute spru\bpe{}d\bpe{}elt in einigen f\bpe{}la\bpe{}chen Se\bpe{}en in Al\bpe{}as\bpe{}ka Me\bpe{}than von selbst \eos{} \\
80 & 22 & Sch\bpe{}on heute spru\bpe{}d\bpe{}elt in einigen f\bpe{}la\bpe{}chen Se\bpe{}en in Al\bpe{}as\bpe{}ka Me\bpe{}than von selbst aus \eos{} \\
90 & 24 & Sch\bpe{}on heute spru\bpe{}d\bpe{}elt in einigen f\bpe{}la\bpe{}chen Se\bpe{}en in Al\bpe{}as\bpe{}ka Me\bpe{}than von selbst aus dem Wasser \eos{} \\
100 & 25 & Sch\bpe{}on heute spru\bpe{}d\bpe{}elt in einigen f\bpe{}la\bpe{}chen Se\bpe{}en in Al\bpe{}as\bpe{}ka Me\bpe{}than von selbst aus dem Wasser . \eos{} \\
\bottomrule
\end{tabular}
\caption{Prefixes of an example German source sentence, for all values of $s$. Lengths do not include \eos.}
\label{tab:partial}
\end{table*}

\section{Related Work}

Closely related to our work are two recent papers by \citet{stahlberg-etal-2022-uncertainty} and \citet{wiher-etal-2022-decoding}, which also explore how degenerate phenomena differ across tasks.
\citet{stahlberg-etal-2022-uncertainty} study two more-constrained tasks, MT and GEC\@. Using exact search and beam search, they find that, for GEC, the distribution is peaked around a few very high-probability outputs, and that, unlike MT, it does not suffer from inadequacy of the mode.

\citet{wiher-etal-2022-decoding} study tasks in the same constrainedness range as we do, from MT to \ucg{}. Although their main focus is on evaluating different decoding strategies (where they confirm the trend seen in the literature, that more constrained tasks favor mode-seeking strategies, while less constrained tasks favor sampling-based methods), they look, as we do, at how degenerate repetition and length bias differ across tasks, finding that these phenomena vary across tasks and decoding methods.

Our contribution here is to provide a more rigorous empirical analysis of why these behaviors differ across tasks. Both \citet{stahlberg-etal-2022-uncertainty} and \citet{wiher-etal-2022-decoding} attribute the differences they observe to task constrainedness, and \citet{stahlberg-etal-2022-uncertainty} quantify task constrainedness by looking at how much the references differ across a multi-reference test set, but neither is able to directly control task constrainedness while keeping all else fixed. Tor our knowledge, our method is the first to study the effect of task constrainedness on degeneration in a completely controlled way.

\section{An Experimental Framework for Controlling Task Constrainedness}
\label{sec:experiment}

The tasks which have been compared before (GEC, MT, story generation, and others) all differ along multiple dimensions besides constrainedness: they use different architectures, different numbers of parameters and amounts of training data, and they produce different length outputs (one sentence for MT, many sentences for story generation), among other distinctions. This makes it difficult to study whether task constrainedness is actually responsible for the differences observed between these tasks.
We therefore seek a way of controlling the constrainedness directly, via some sort of ``knob'' that we could adjust. In this section, we introduce an experimental framework that allows us to do so.

\subsection{Truncation}

We begin with an ordinary MT dataset and a desired constrainedness level $s$, which can be $0$ (\ucg{}, the least constrained task) or $100$ (MT, the most constrained task in our setup) or anything in between.
In our experiments, we choose $s = 0, 10, \ldots, 100$. For each value of $s$, we truncate each source sentence in the dataset to $s\%$ of its original length. To be precise, if $x = x_1 \cdots x_n \cdot \eos$ is the original sentence (after separating punctuation, but before BPE), we let $n' = \lceil n \cdot s\%\rceil$ and truncate the sentence to $x_1 \cdots x_{n'} \cdot \eos$. See \cref{tab:partial} for an example German source sentence and all of its truncations.

We apply this truncation to all of the source sentences in the train, dev, and test data, leaving the target sentences intact. This way, as $s$ decreases, the model has to predict the target side given less and less information about what it might contain. Or, to think of it another way, as $s$ decreases, there become more and more possible ``correct'' answers, since the truncated source sentence could be the prefix of many possible full source sentences, and a translation of any one of them can be considered a valid solution to the task.

\subsection{Experimental details}
\label{subsec:experimental-details}

\begin{table} \centering 
\begin{tabular}{lrr}
\toprule
& de-en & zh-en \\
\midrule
train & 205,898 & 231,259 \\
dev   & 888     & 879 \\
test  & 8,079   & 8,549 \\
\bottomrule
\end{tabular}
\caption{Sizes of our training, development, and test datasets.}
\label{tab:data}
\end{table}

We use the German-to-English (de-en) and Chinese-to-English (zh-en) datasets from IWSLT 2017 \citep{cettolo-etal-2012-wit3}, consisting of transcribed TED talks. We use the standard dataset for training, the 2010 development set for development, and the 2010--2015 test sets for testing, following the split by \citet{kulikov-etal-2021-characterizing}. Table~\ref{tab:data} shows the size of each of these sets, after removing copy noise (pairs where the source and target are identical) from the data \citep{ott-etal-2018-analyzing}.

We preprocess the data using BPE tokenization \citep{sennrich-etal-2016-neural}. To ensure that the experimental setup is as similar as possible for all values of $s$, we learn BPE on the full, untruncated dataset. Then, once BPE has been learned, we apply it to the truncated data. Initially, we experimented with both joint and separate BPE, but found very little difference between them, so we present results for joint BPE only.

For our MT system, we use the Transformer model \citep{NIPS2017_3f5ee243}; specifically, we use a fork of the Transformers without Tears library \citep{nguyen-salazar-2019-transformers}.\footnote{\url{https://github.com/darcey/transformers_without_tears/tree/mt-interpolation-paper}}. We use identical hyperparameter settings for both language pairs and all values of $s$; these are the same as the Transformers without Tears base configuration, except that we use 6 layers and 4 heads. 

We trained our systems both with and without label smoothing \citep{szegedy+:2016}, thinking that, because label smoothing changes the shape of the distribution, it might impact the results. We discuss the effect of label smoothing in \S\ref{sec:label-smoothing}; in all other sections we look only at systems trained without it. All of our results are averaged across three random restarts.

Fearing that BLEU scores might not provide a meaningful enough signal for $s < 100$, we tried using both dev BLEU and dev perplexity to lower the learning rate and control early stopping; these gave very similar results, so we only present results for the systems tuned using dev BLEU\@. 

To better view the natural properties of the distribution, we do not use any length normalization during decoding. We decode up to a maximum length of 300 tokens.

We make our full experimental setup publicly available on GitHub.\footnote{\url{https://github.com/darcey/mt-interpolation}}

\subsection{Sanity checks}
\label{subsec:peakedness}

\begin{figure*}
     \centering
     \begin{subfigure}[b]{0.31\textwidth}
         \includegraphics[width=\textwidth,trim=15 15 15 15]{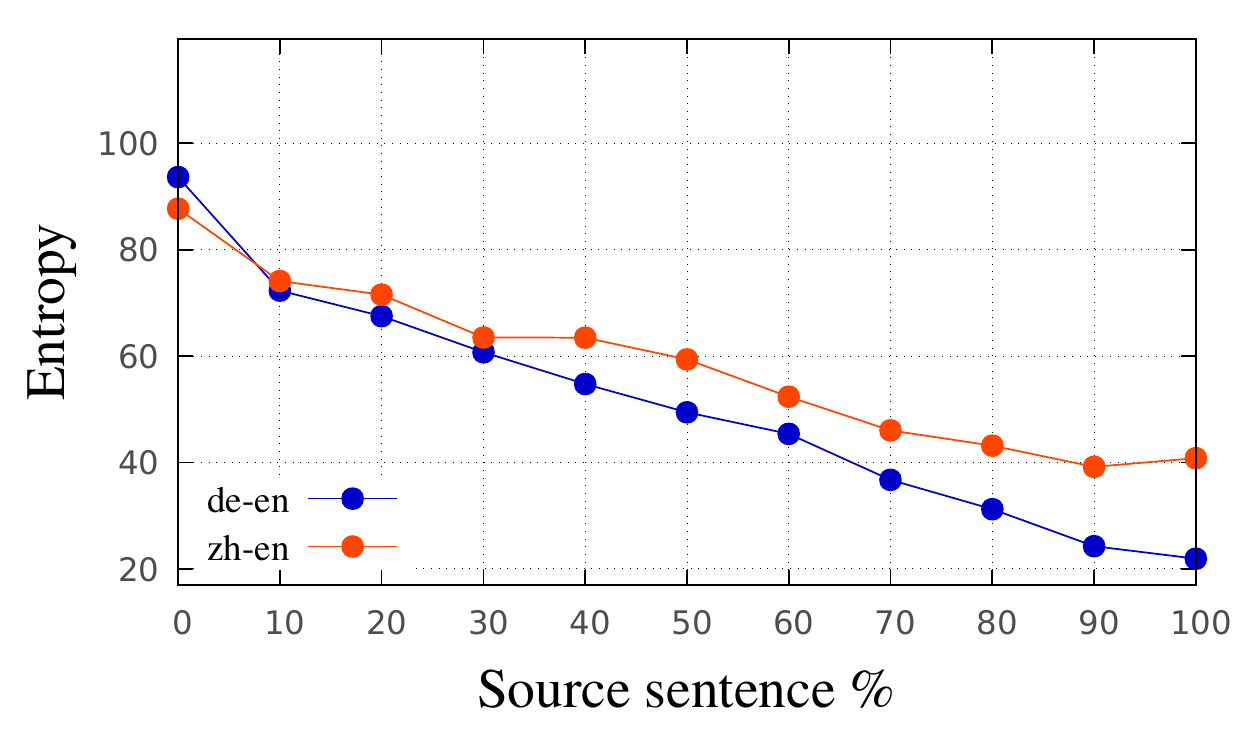}
         \caption{Per-sentence entropy (nats)}
         \label{fig:peakedness_entropy}
     \end{subfigure}%
     \hfill
     \begin{subfigure}[b]{0.31\textwidth}
         \includegraphics[width=\textwidth,trim=15 15 15 15]{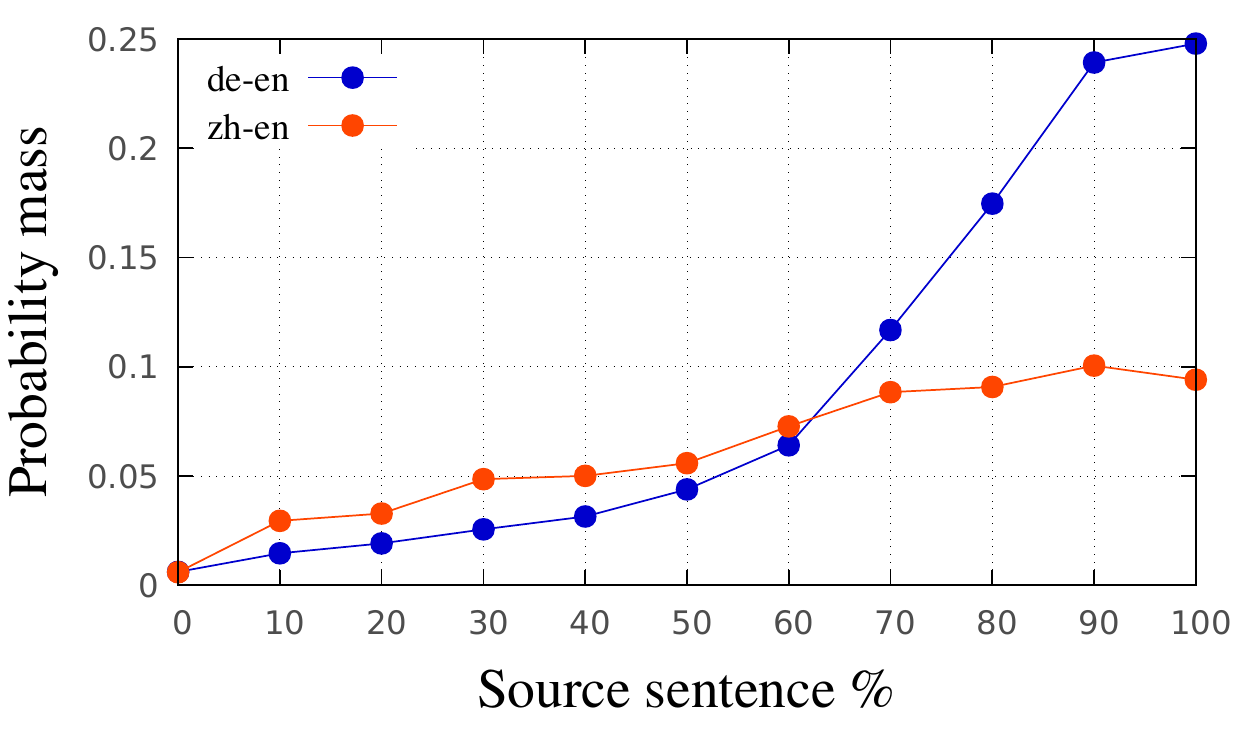}
         \caption{Total probability mass}
         \label{fig:peakedness_mass}
     \end{subfigure}%
     \hfill
     \begin{subfigure}[b]{0.31\textwidth}
         \includegraphics[width=\textwidth,trim=15 15 15 15]{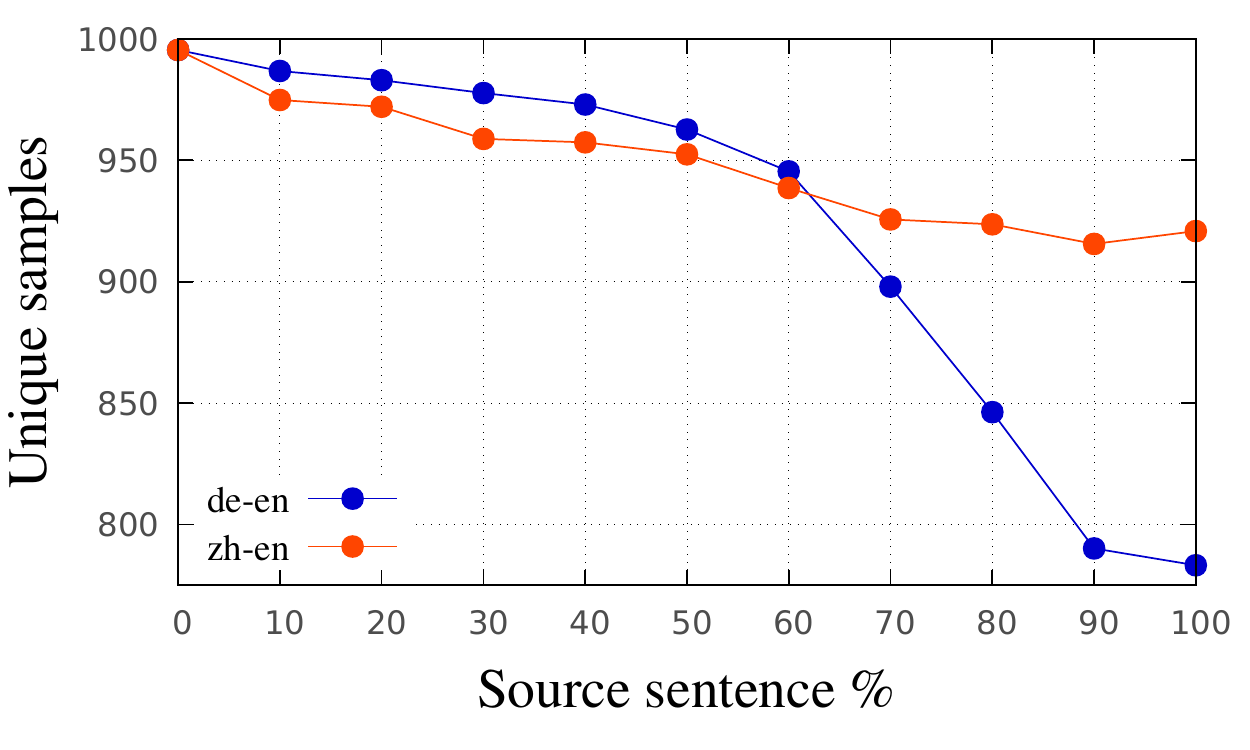}
         \caption{Number of unique samples}
         \label{fig:peakedness_unique}
     \end{subfigure}%
     \caption{Increasing constrainedness increases the peakedness of the predictive distribution as expected. Every data point is based on 1000 random samples for each sentence in the test data.}
     \label{fig:peakedness}
\end{figure*}

\begin{figure}[!ht]
\centering
    \includegraphics[width=0.45\textwidth,trim=15 15 10 10]{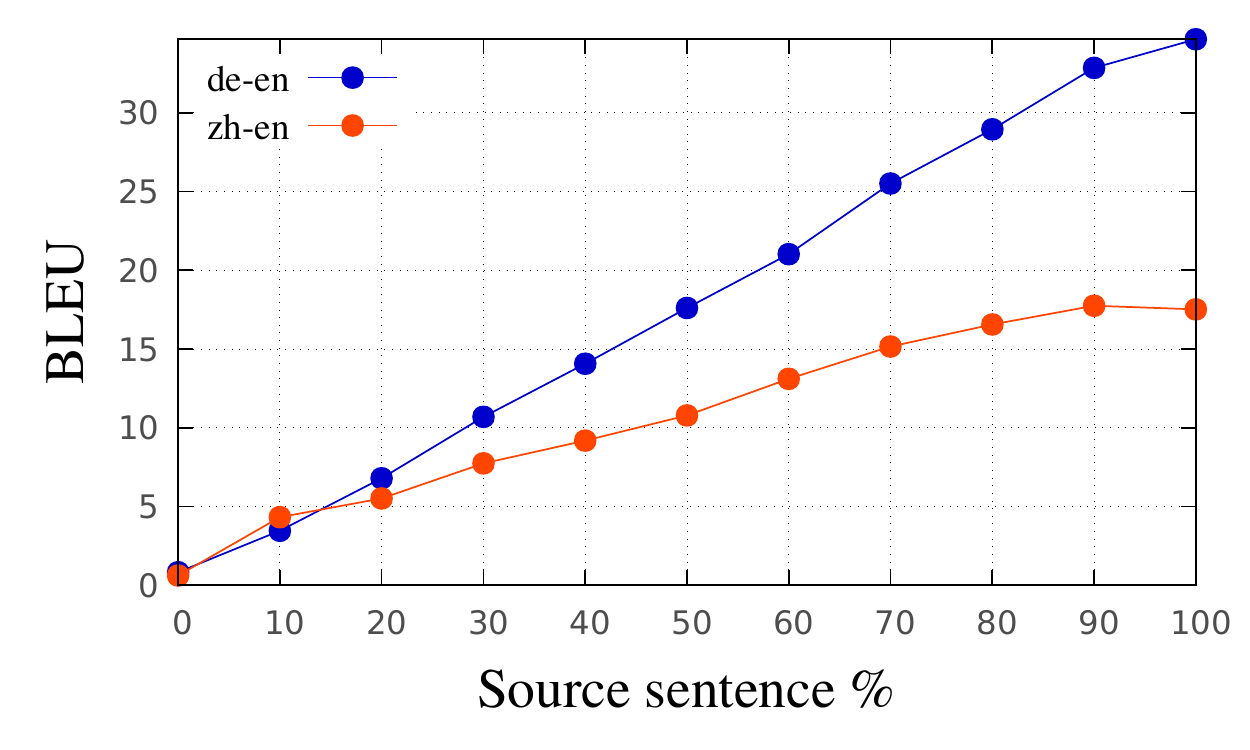}
    \caption{Predictably, BLEU score decreases smoothly as we decrease $s$.}
\label{fig:bleu}
\end{figure}

We verify via BLEU score that we have trained our systems successfully. Although, for the purposes of our experiment, it is not necessary to use a state-of-the-art MT system, we nonetheless achieve reasonable BLEU scores of 34.7 and 17.5 for de-en and zh-en respectively for the $s=100$ systems, using the standard beam size of 4 for decoding. Predictably, lowering $s$ also decreases the BLEU score, as can be seen in Figure~\ref{fig:bleu}.

As an additional sanity check, we confirm that varying $s$ does indeed change the spread of the distribution in the expected way. Using 1000 samples for each sentence in the test set, we estimate the entropy (Figure \ref{fig:peakedness_entropy}), and find that it decreases as $s$ increases. In addition, following \citet{ott-etal-2018-analyzing}, we look at the portion of the total probability mass covered by all of the unique samples, and find that, although the number of unique samples decreases as $s$ increases (Figure~\ref{fig:peakedness_unique}), the total probability mass covered increases (Figure~\ref{fig:peakedness_mass}).

\section{Degeneracy in the Mode}

In this section, we look at how length bias and repetition vary as we vary the constrainedness parameter $s$.

\subsection{Length bias}

\begin{figure}[!ht]
\centering
    \begin{subfigure}[b]{0.45\textwidth}
    \includegraphics[width=\textwidth,trim=15 15 10 10]{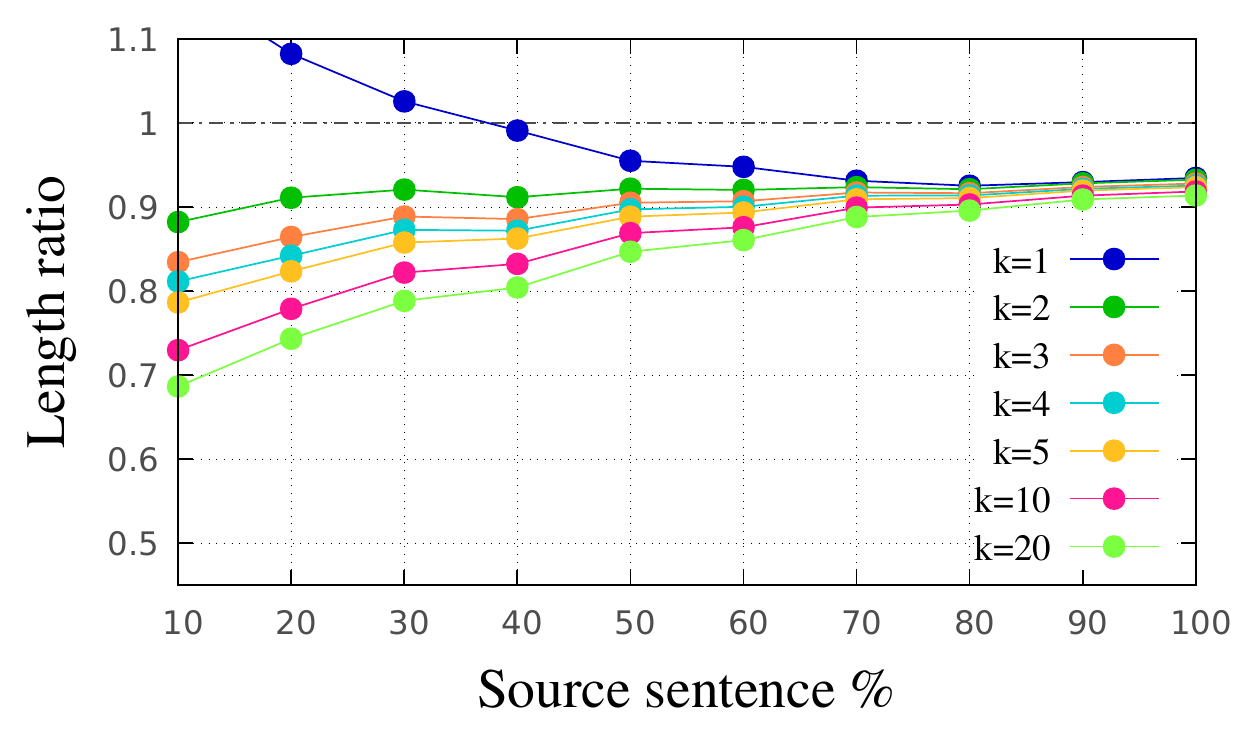}
    \caption{German--English}
    \end{subfigure}
    \\
    \vspace{3ex}
    \begin{subfigure}[b]{0.45\textwidth}
    \includegraphics[width=\textwidth,trim=15 15 10 10]{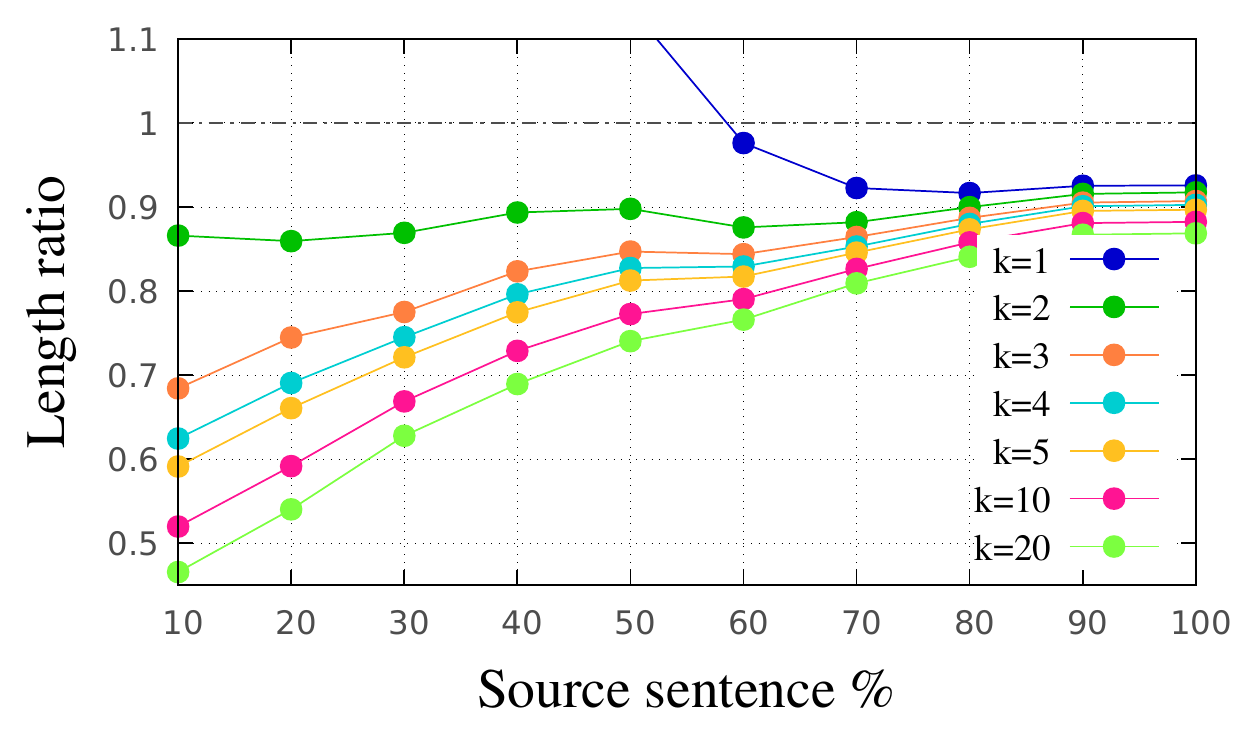}
    \caption{Chinese--English}
    \end{subfigure}
\caption{Length ratio versus source sentence percentage ($s$), for various beam sizes ($k$). For high $s$, there is a slight bias towards shorter outputs that increases mildly with $k$, whereas for low $s$, we see extreme bias, towards longer or shorter outputs depending on $k$.}
\label{fig:length}
\end{figure}

Length bias is a problem where the length of the output consistently differs from the length of the reference; the term typically refers to sentences being too short. In NMT, length bias is such a major and well-known problem that nearly all systems correct for it using some kind of length normalization during decoding \citep{wu+:2016,koehn-knowles-2017-six,murray-chiang-2018-correcting}.

In NMT, length bias gets worse the closer one approaches the mode of the distribution. It has been repeatedly shown that, as beam size increases, bringing the output translation closer to the mode, the length bias becomes more extreme. In fact, the mode of the distribution is often simply the empty string itself \citep{stahlberg-byrne-2019-nmt}.

On the other hand, length bias has been under-studied in less constrained tasks such as story generation or \ucg{}. We know of just two reports of this problem: for story generation, \citet{holtzman-etal-2020-curious} report worsening length bias as beam size increases, with immediate stopping when using beam sizes $\geq 64$, and \citet{wiher-etal-2022-decoding} found length bias for beam sizes $k=5,10$; however, neither of these is the main result of their respective papers.

This difference in emphasis seen in the literature would seem to suggest that length bias only affects MT, and does not affect less constrained tasks like story generation. Thus, if constrainedness were fully responsible for the difference seen in the literature, then we would expect to see length bias decrease with constrainedness, becoming less of a problem for less constrained tasks.

To test this, we measure how length bias changes as we vary~$s$. To quantify length bias, we compute the (micro-averaged) \textit{length ratio},
\begin{equation*}
    \ell(T) = \frac{\sum_{(h,r) \in T} |h|}{\sum_{(h,r) \in T} |r|}
\end{equation*}
where $T$ is a test set consisting of pairs $(h, r)$, where $h$ is a hypothesis (output) sentence and $r$ is a reference sentence.

Figure \ref{fig:length} shows how length ratio changes as we vary both $s$ and the beam size $k$.\footnote{These graphs (and all of our beam search results) exclude the $s=0$ case, which turn out to be nearly meaningless: since the source sentence is identical (namely, the empty string) for each sentence pair in the test set, the beam search results will also be identical, meaning that the decoder will simply generate $|T|$ copies of the same sentence. So any properties of that one sentence will be magnified. The results under beam search for $s=0$ are therefore little better than noise.}
Consistent with previous findings, we find that standard NMT suffers from considerable length bias, with the problem worsening as beam size $k$ increases. But to our surprise, as $s$ decreases, we find that not only does length bias worsen, but that the dependence on beam size grows stronger and stronger. This is surprising given the lack of concern with length bias in the literature on less constrained tasks. To our knowledge, we are the first to report a result like this, where length bias actually worsens as task constrainedness decreases.

We can think of two explanations for this result. The first is that there are other factors besides constrainedness affecting the length bias seen across tasks. We suspect that the length of the reference outputs might be a major part of this; models like GPT-2 are trained to produce much larger chunks of text than our systems, which typically just output one sentence at a time. The second is that this is an artifact of the decoding processes used. Most of the literature on NMT uses mode-seeking decoding strategies such as beam search, while literature on less constrained generation favors sampling-based approaches. So it could in fact be that all unconstrained systems also suffer from length bias, but it simply doesn't show up because beam search is not used with those systems.
We also note that it may be more difficult to study length bias in less constrained tasks, since there is not necessarily a roughly ``correct'' length the way there is in MT\@.

A last interesting result is that, for $k=1$ (greedy search), the length ratio actually increases for decreasing $s$, ending up well above 1. This agrees with a recent result reported by \citet{wiher-etal-2022-decoding}, who found that, for the relatively unconstrained task of story generation, beam sizes $k=5,10$ returned texts that were too short, while greedy search returned texts which were far too long.

\subsection{Repetition}

\begin{figure*}
\centering
    \begin{subfigure}[b]{0.45\textwidth}
        \includegraphics[width=\textwidth,trim=15 15 10 10]{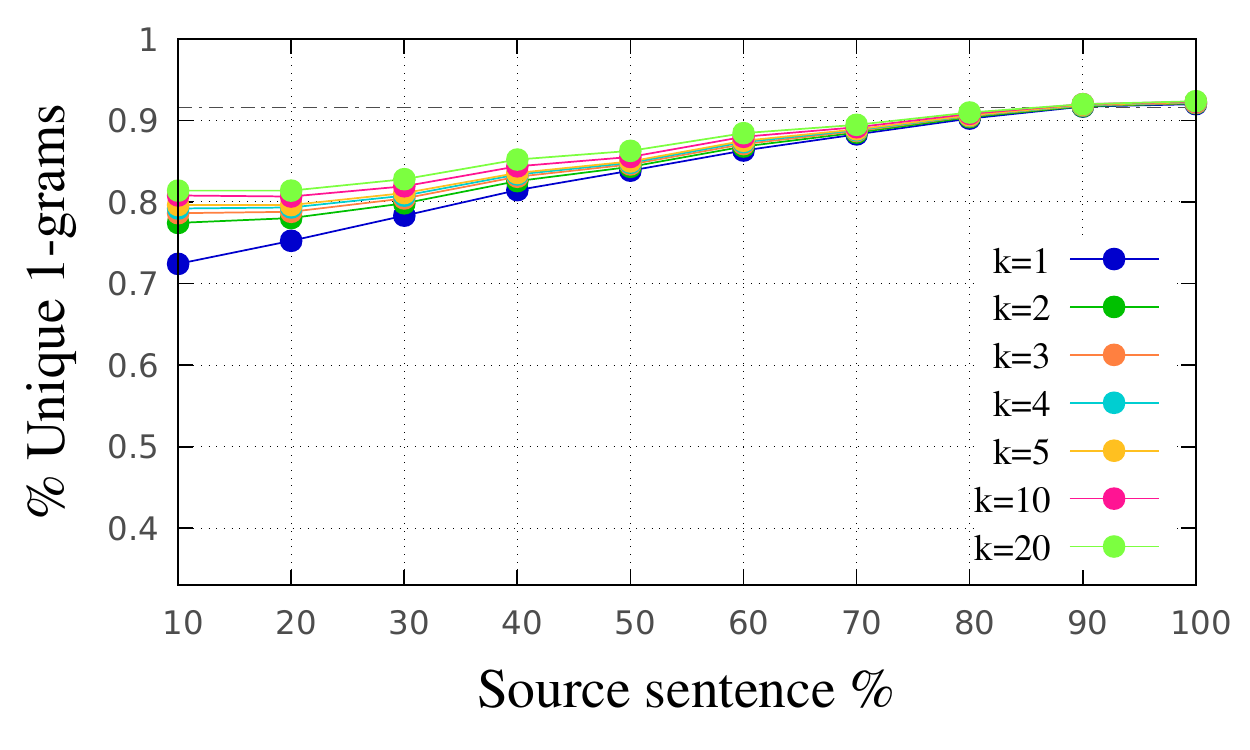}
    \end{subfigure}%
    \hfill
    \begin{subfigure}[b]{0.45\textwidth}
        \includegraphics[width=\textwidth,trim=15 15 10 10]{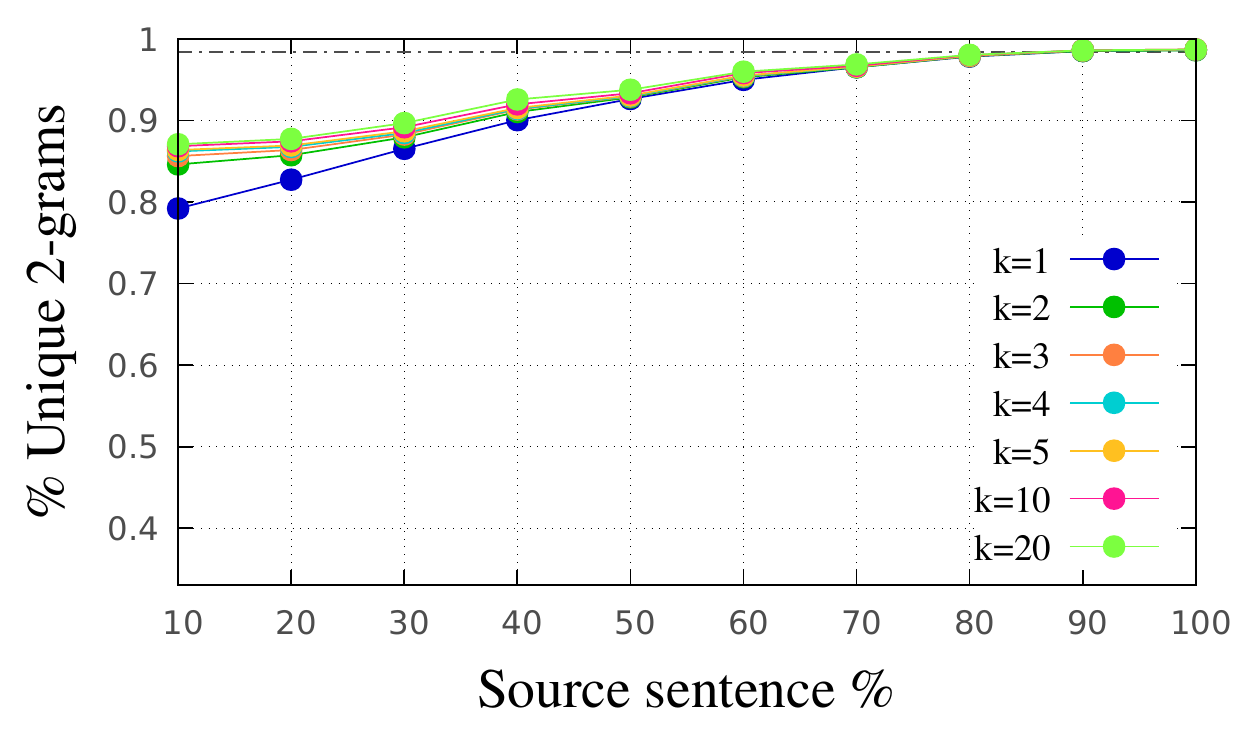}
    \end{subfigure}
    \\ \vspace{-0.73cm}
    \vspace{3ex}
    \begin{subfigure}[b]{0.45\textwidth}
        \begin{mdframed}[backgroundcolor=white,linewidth=0pt,innertopmargin=0pt,innerbottommargin=0pt,innerleftmargin=0pt,innerrightmargin=0pt]
        \includegraphics[width=\textwidth,trim=15 15 10 10]{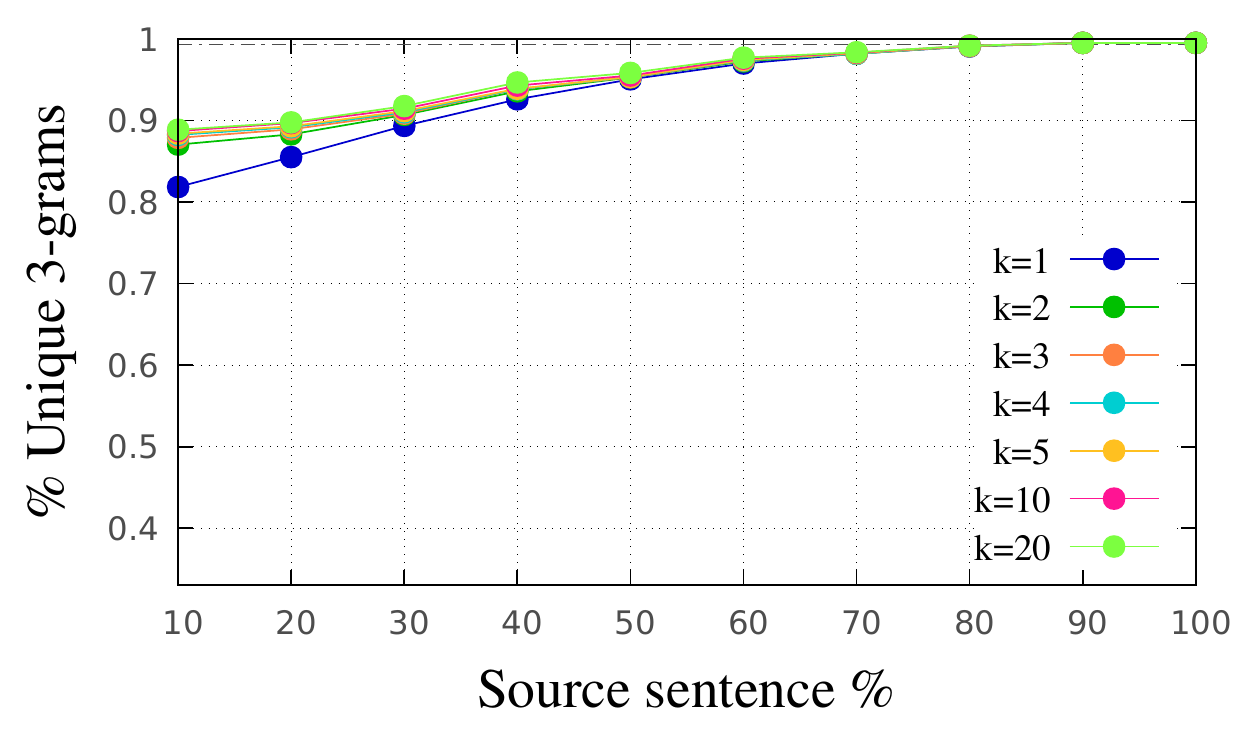}
        \end{mdframed}
    \end{subfigure}%
    \hfill
    \begin{subfigure}[b]{0.45\textwidth}
        \begin{mdframed}[backgroundcolor=white,linewidth=0pt,innertopmargin=0pt,innerbottommargin=0pt,innerleftmargin=0pt,innerrightmargin=0pt]
        \includegraphics[width=\textwidth,trim=15 15 10 10]{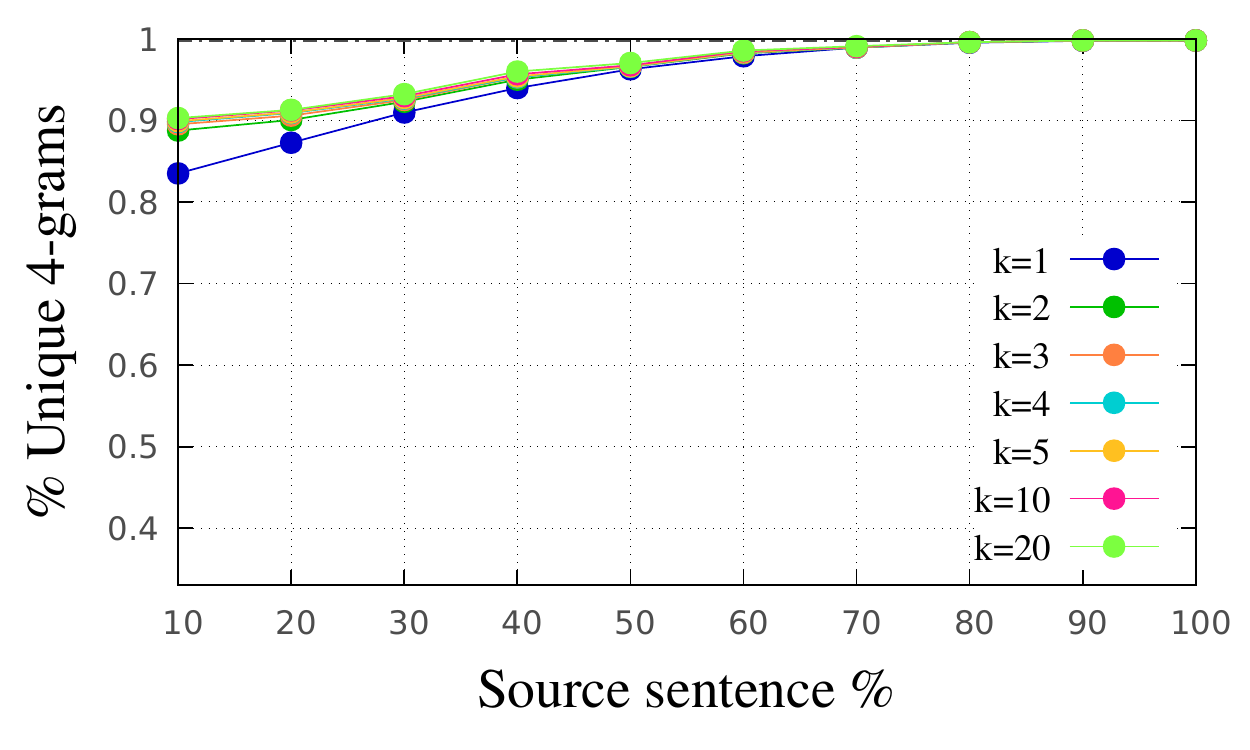}
        \end{mdframed}
    \end{subfigure}
    \caption{Amount of repetition versus source sentence percentage ($s$), for various beam sizes ($k$). Repetition is measured as the percentage of unique $n$-grams in a sentence; the graphs show this for different values of $n$. The repetition rate of the reference is plotted as a dashed grey line. Across all values of $n$, the percent of unique $n$-grams drops as $s$ decreases. These graphs show German-to-English results only; see Appendix~\ref{appendix:examples} for Chinese-to-English.}
    \label{fig:repetition-de}
\end{figure*}

Degenerate repetition is a well-known problem where the model gets stuck in a loop, repeating the same $n$-grams over and over again. It so strongly affects less constrained tasks like story generation \citep{holtzman-etal-2020-curious} that these tasks avoid mode-seeking strategies altogether, preferring sampling-based approaches. Since pure random sampling also produces low-quality output, most work on this topic has focused on finding a balance between mode-seeking and pure sampling, either by truncating the distribution during sampling \citep{fan-etal-2018-hierarchical,holtzman-etal-2020-curious,basu2021mirostat,zhang-etal-2021-trading,nadeem-etal-2020-systematic,delucia-etal-2021-decoding}, or by using some combination of sampling and search \citep{massarelli-etal-2020-decoding}, though \citet{welleck-etal-2020-neural} address the issue via training rather than search, by modifying the objective function to discourage repetition.

In contrast, to our knowledge, degenerate repetition has not been reported in the literature on NMT. (Our anecdotal experience is that degenerate repetition is a familiar sight in MT, but not a serious problem in well-trained systems.) If the difference between story generation and MT can be explained by their constrainedness, as previous work has suggested, then we should expect to see repetition increase smoothly as we decrease $s$.

This is, in fact, exactly what we find. Figure~\ref{fig:repetition-de} shows the amount of repetition for German-to-English, measured as the percentage of unique $n$-grams which appear in each search result (that is, for each search result, the number of $n$-gram types divided by tokens), as compared to the reference. (The Chinese-to-English results are similar, and can be found in Appendix~\ref{appendix:examples}.)
We find that, as $s$ decreases, repetition increases considerably. Consistent with the literature, we see basically no evidence of repetition in pure MT, where the amount of repetition almost perfectly matches that seen in the references. But as $s$ decreases, so does the percent of unique $n$-grams, until for $s=10$ there is very clear evidence of repetition. We therefore feel confident in concluding that task constrainedness adequately explains the difference in the level of concern paid to repetition in the literature for different tasks.

One interesting thing to observe is that, as beam size increases,  repetition actually decreases. We suspect that this might be due to the effect of length bias: as shown in the previous section, higher beam sizes tend to return shorter sentences, and these seem less likely to experience degenerate repetition (though it is certainly possible to have both problems at once).

\subsection{Discussion}

\begin{table*}[!ht] \centering \scriptsize
\begin{tabular}{@{}rl@{}}
\toprule
$s$ (\%) & Output found using beam search, $k=4$ \\
\midrule
0 & And I said, ``Well, I'm going to show you a little bit.'' \\
10 & When Steve Lopez said, ``You know, I'm not going to be here.'' \\
20 & When Steve Lopez, Columni, who is the first person in the world, he's the first person in the world, and he's the first person in the world. \\
30 & When Steve Lopez, Columni, the Los Angeles Times, he said, ``You know, I'm going to go to school.'' \\
40 & When Steve Lopez, Columnist, the Los Angeles Times, one day, he said, ``You know, we're going to have to do this.'' \\
50 & When Steve Lopez, Columnist, the Los Angeles Times, one day through the Pacific Ocean Ocean, I started to think about it. \\
60 & When Steve Lopez, Columnist in Los Angeles Times, one day through the streets in the center of the city, the city of New York. \\
70 & When Steve Lopez, Columnist, the Los Angeles Times, one day through the streets of Los Angeles, the city of London. \\
80 & When Steve Lopez, Columnist, the Los Angeles Times, one day went through the streets at the center of Los Angeles, I heard this story. \\
90 & When Steve Lopez, Columnist, the Los Angeles Times, one day went through the streets at the center of Los Angeles, he heard a wonderful story. \\
100 & When Steve Lopez, Columnist at the Los Angeles Times, walked through the streets at the center of Los Angeles, he heard a wonderful music. \\
\midrule
ref & One day, Los Angeles Times columnist Steve Lopez was walking along the streets of downtown Los Angeles when he heard beautiful music. \\
\bottomrule
\end{tabular}
\caption{Beam search ($k=4$) outputs for a sentence in the test dataset, shown across all values of $s$. Illustrates both length bias and degenerate repetition.}
\label{tab:ex1}
\end{table*}

It is illustrative to examine some strings generated by our systems. Table~\ref{tab:ex1} shows the translations for one sentence from the test data; others can be found in Appendix~\ref{appendix:examples}. Consistent with our results, we see length roughly decrease along with $s$. We also see some concrete examples of degenerate repetition for the lower $s$ values. As is typical, the same phrase is repeated over and over, separated by commas or ``and''.

In addition to length bias and repetition, we can also observe that, as $s$ decreases, the content of the generated strings diverges further and further from the reference. But we notice that, qualitatively, as it does this, the outputs get increasingly boring. This fits with what others have reported \citep{holtzman-etal-2020-curious}, that beam search simply produces tedious and boring output for less constrained tasks.

Another observation is that some of these sentences are simply ungrammatical. While grammatical errors are very common among random samples, it is interesting to see them even at these high probabilities.

\section{No Degeneracy in Samples}

\begin{figure}[!ht]
\centering
    \includegraphics[width=0.45\textwidth,trim=15 15 10 10]{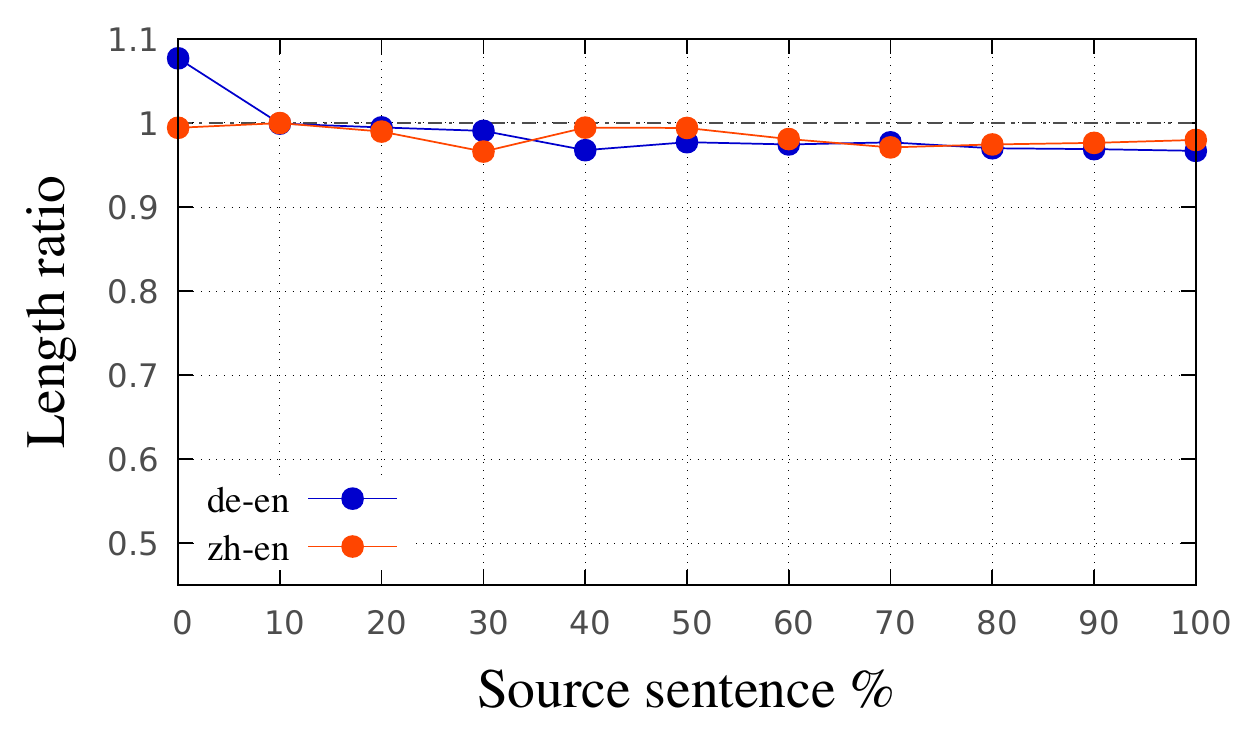}
    \caption{Length ratio of samples versus source sentence percentage ($s$), for both language pairs. The samples only suffer from very slight length bias, and only for higher values of $s$.}
\label{fig:length-samples}
\end{figure}

\begin{figure*}
\centering
    \begin{subfigure}[b]{0.45\textwidth}
        \includegraphics[width=\textwidth,trim=15 15 10 10]{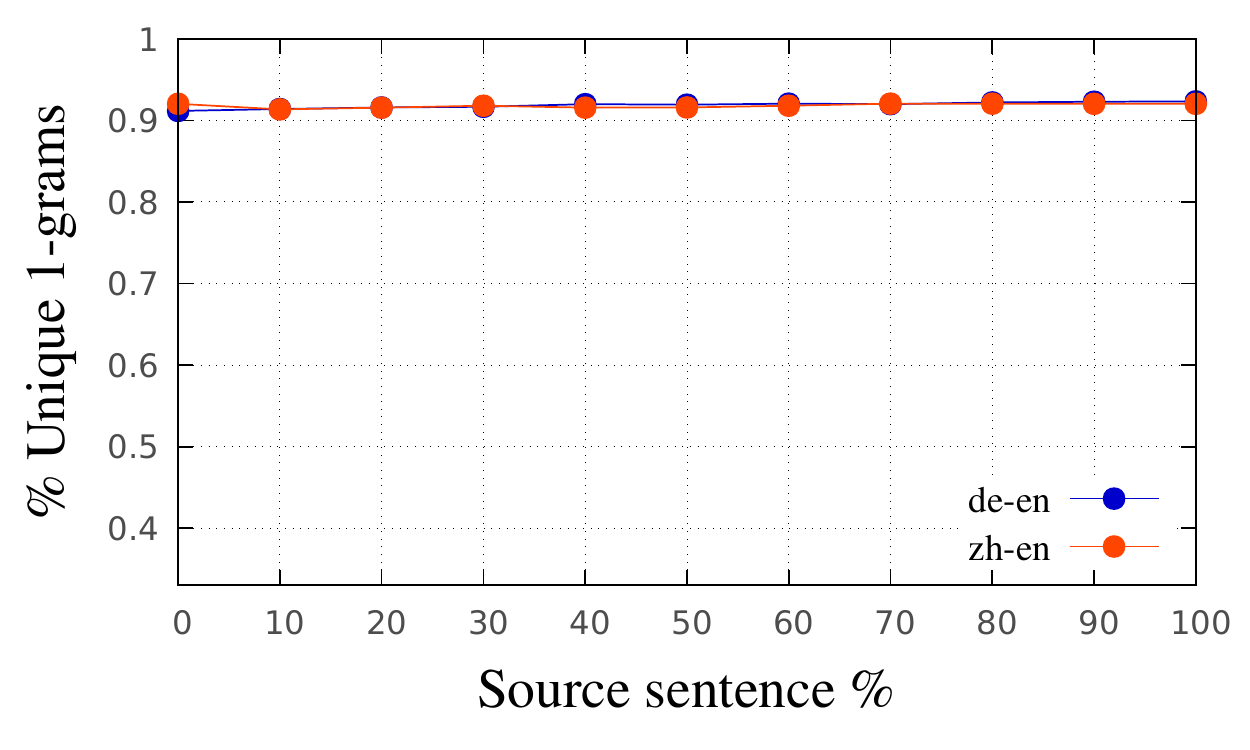}
    \end{subfigure}%
    \hfill
    \begin{subfigure}[b]{0.45\textwidth}
        \includegraphics[width=\textwidth,trim=15 15 10 10]{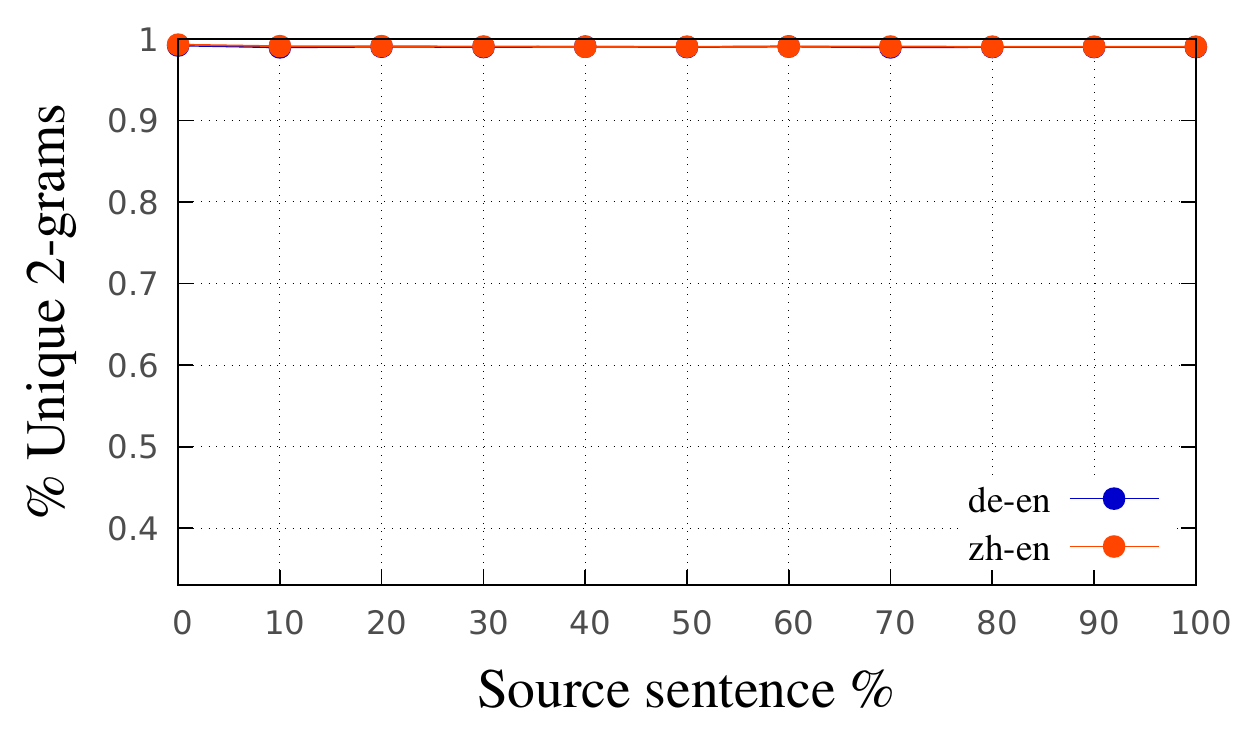}
    \end{subfigure}
    \\ \vspace{-0.73cm}
    \vspace{3ex}
    \begin{subfigure}[b]{0.45\textwidth}
        \begin{mdframed}[backgroundcolor=white,linewidth=0pt,innertopmargin=0pt,innerbottommargin=0pt,innerleftmargin=0pt,innerrightmargin=0pt]
        \includegraphics[width=\textwidth,trim=15 15 10 10]{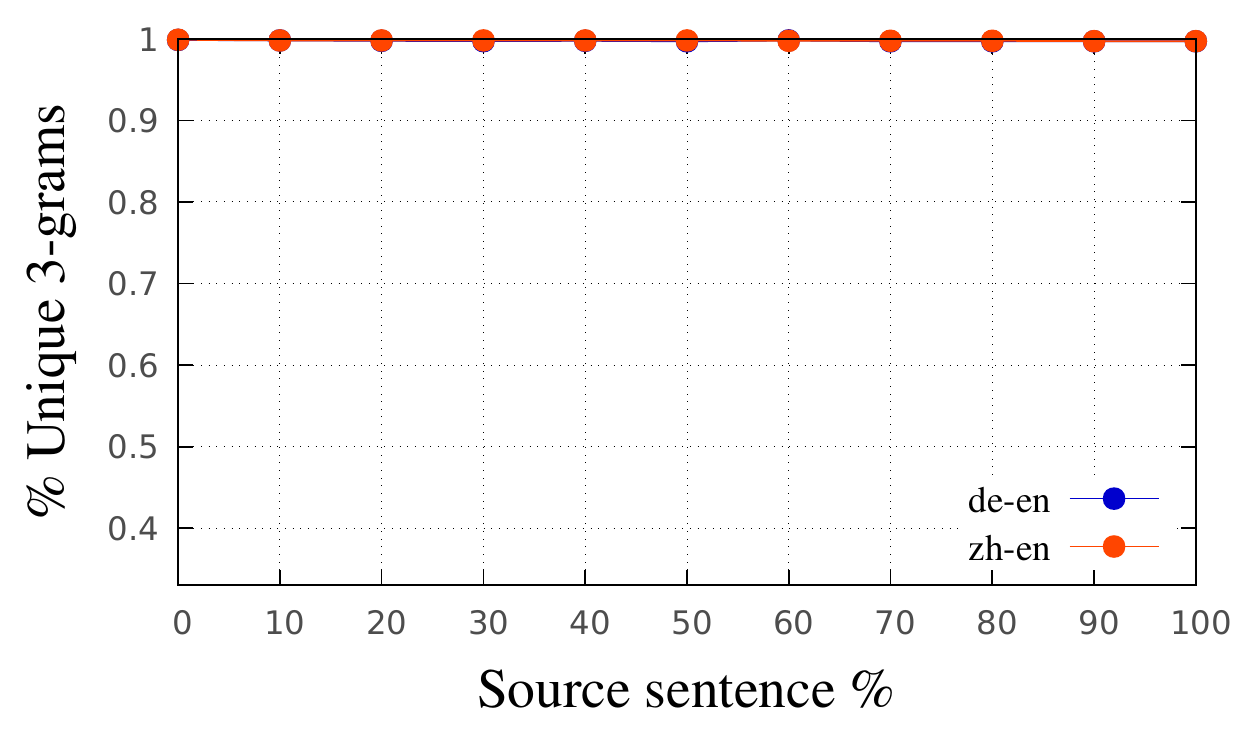}
        \end{mdframed}
    \end{subfigure}%
    \hfill
    \begin{subfigure}[b]{0.45\textwidth}
        \begin{mdframed}[backgroundcolor=white,linewidth=0pt,innertopmargin=0pt,innerbottommargin=0pt,innerleftmargin=0pt,innerrightmargin=0pt]
        \includegraphics[width=\textwidth,trim=15 15 10 10]{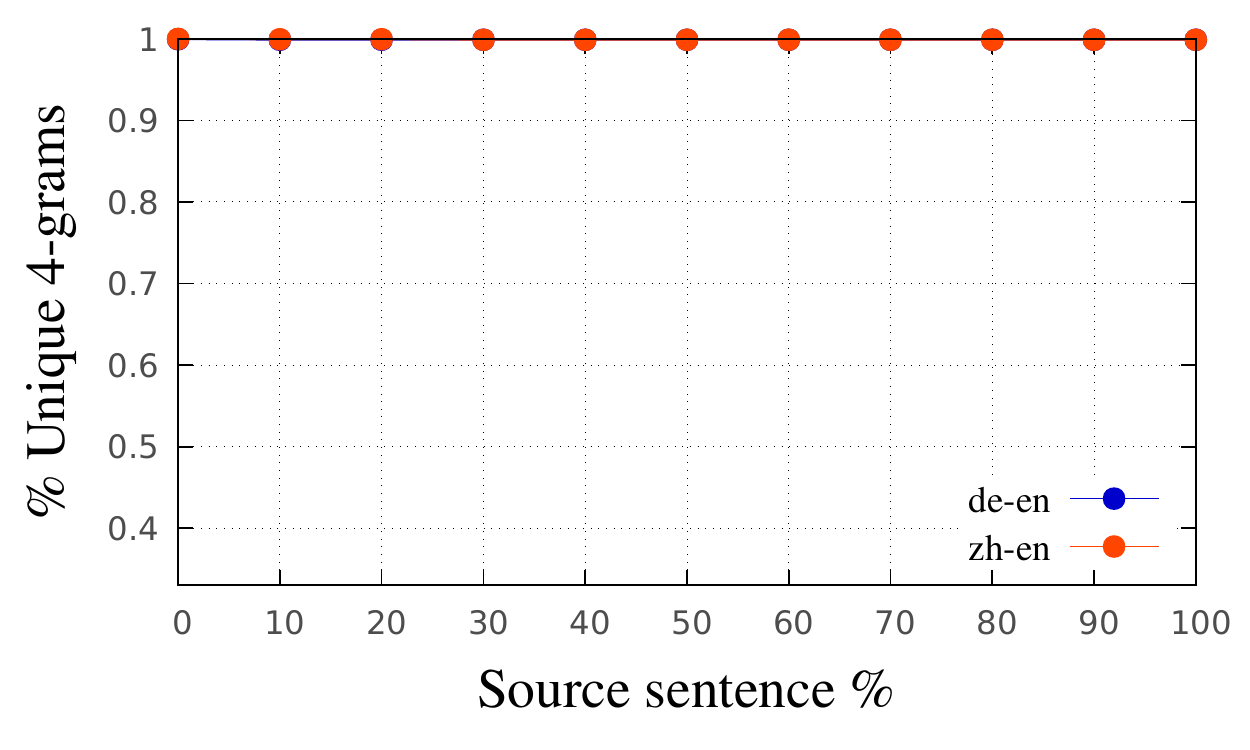}
        \end{mdframed}
    \end{subfigure}
    \caption{Amount of repetition versus source sentence percentage ($s$), for various values of $n$, computed over 1000 random samples for each sentence in the test data. The samples show no evidence of degenerate repetition whatsoever; the level of repetition matches extremely closely to the reference (shown as a dashed grey line which is completely hidden behind the sampling results).}
    \label{fig:repetition-samples}
\end{figure*}

In addition to looking at search results, we also look at samples from the distribution. For each system and for each sentence in the test set, we take 1000 samples, and discover that the samples do not suffer from either degenerate repetition or length bias (\cref{fig:length-samples,fig:repetition-samples}). This underscores that these problems are specific to the mode, and are not properties of the distribution as a whole.

For low values of $s$, this should not be particularly surprising; sampling-based decoding approaches such as top-$k$ \citep{fan-etal-2018-hierarchical} and top-$p$ \citep{holtzman-etal-2020-curious} are favored for these tasks specifically to avoid the degenerate mode.

Yet it may be surprising to see that the pure MT ($s=100$) outputs do not suffer from degeneracy either. Since MT papers rarely explore properties of the full distribution beyond the mode, one might get the false impression that length bias is a problem that affects most sentences in the distribution. Figure~\ref{fig:length-samples} shows that this is definitely not the case. This supports the argument by \citet{eikema-aziz-2020-map} that it is a mistake to focus too much on the mode during decoding.

\begin{figure*}
     \centering
     \begin{subfigure}[b]{0.31\textwidth}
         \includegraphics[width=\textwidth,trim=15 15 15 15]{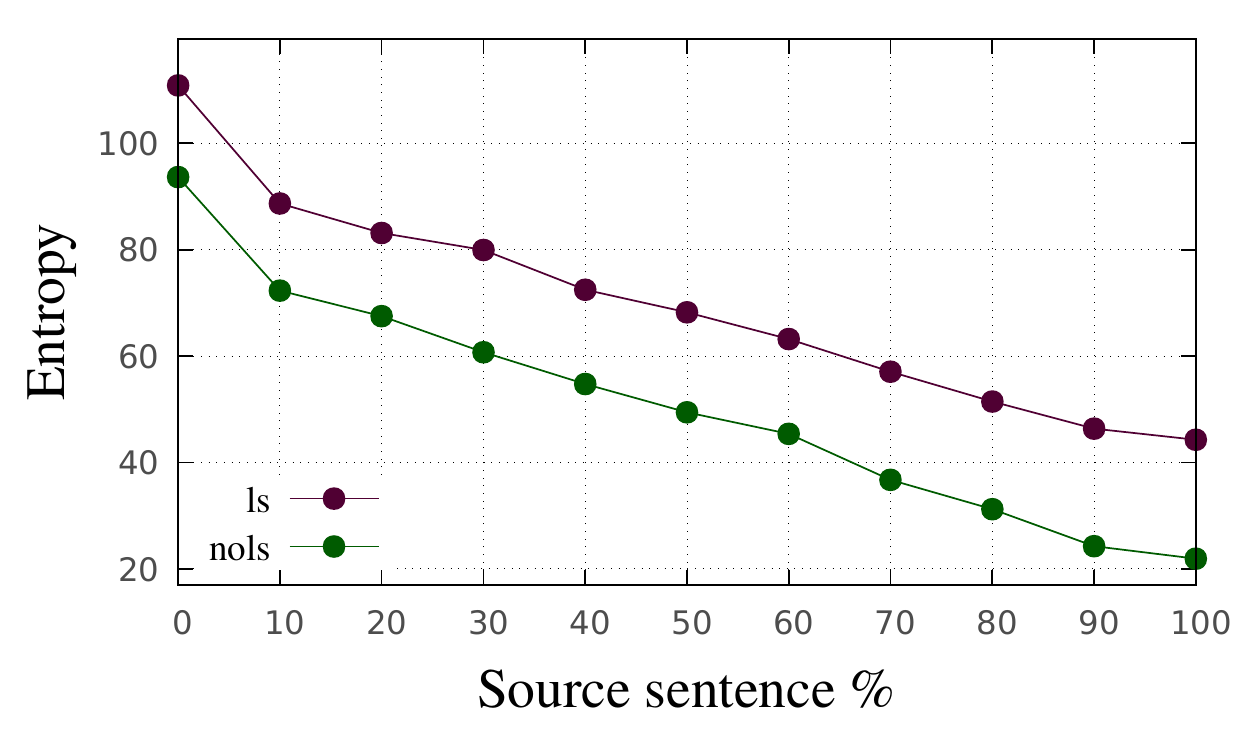}
         \caption{Per-sentence entropy (nats), de-en}
     \end{subfigure}%
     \hfill
     \begin{subfigure}[b]{0.31\textwidth}
         \includegraphics[width=\textwidth,trim=15 15 15 15]{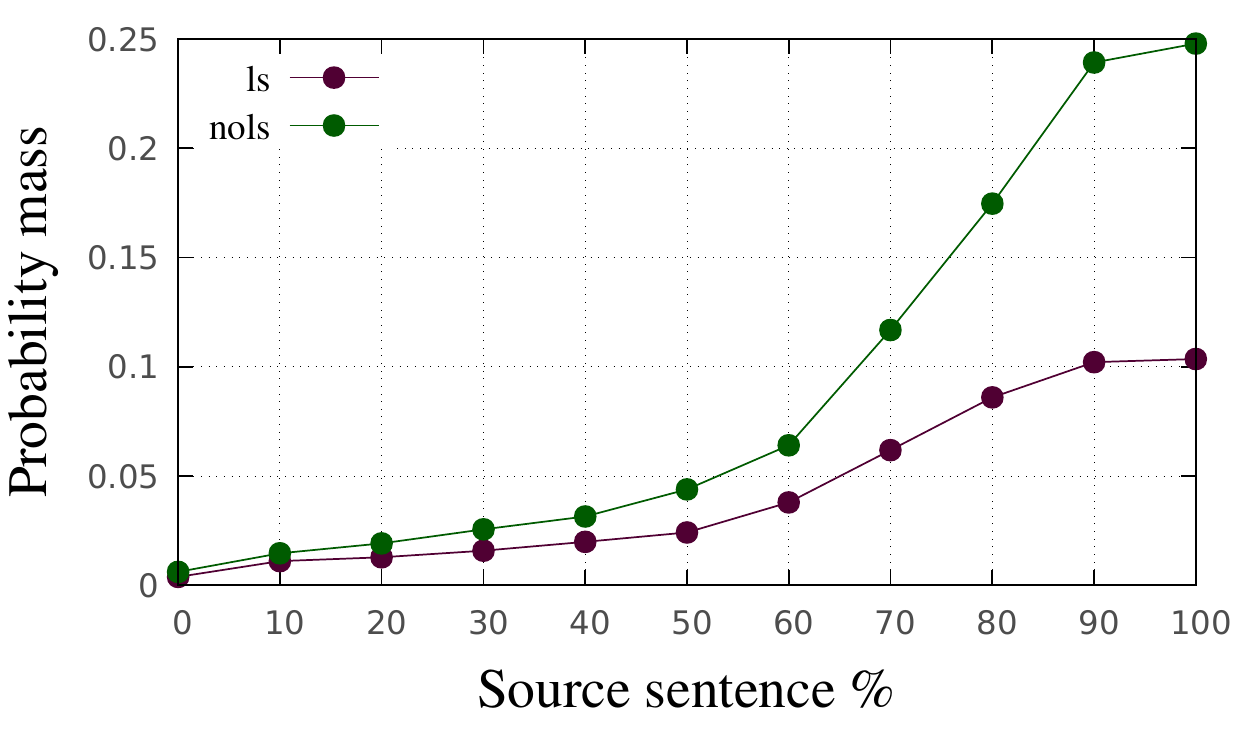}
         \caption{Total probability mass, de-en}
     \end{subfigure}%
     \hfill
     \begin{subfigure}[b]{0.31\textwidth}
         \includegraphics[width=\textwidth,trim=15 15 15 15]{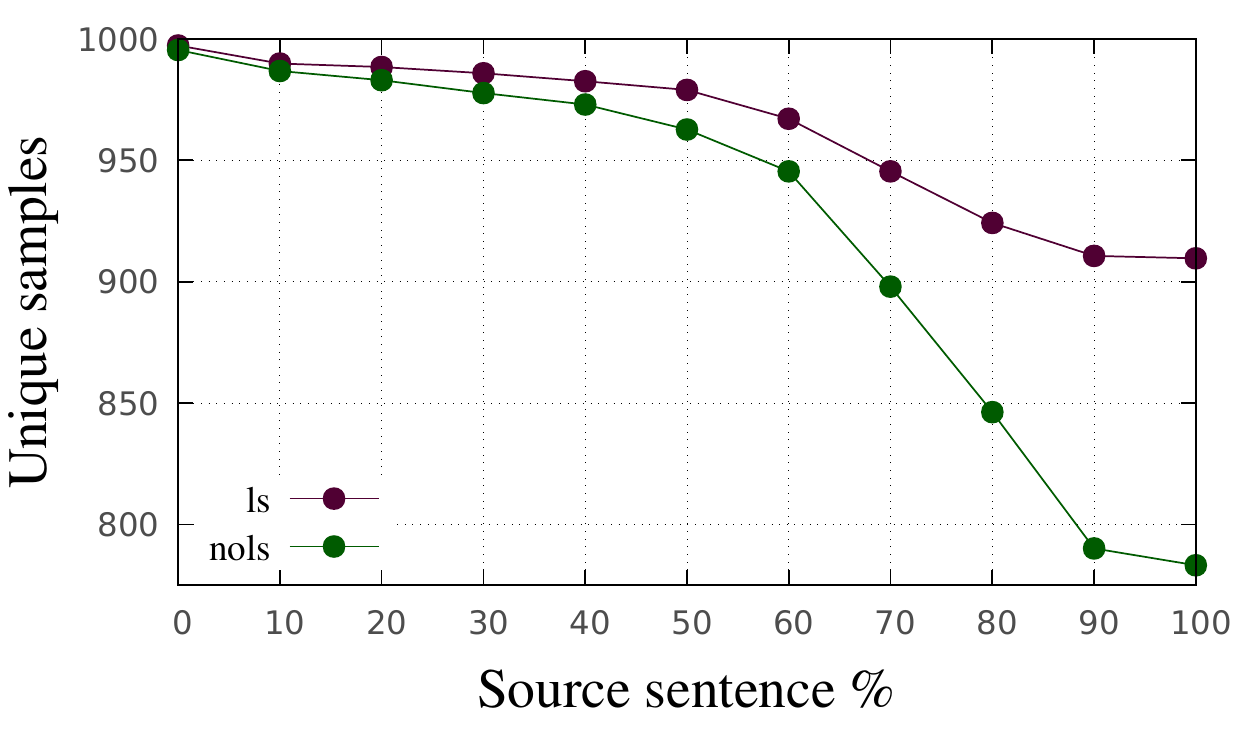}
         \caption{Number of unique samples, de-en}
     \end{subfigure}%
     \caption{Effect of label smoothing (ls) on the peakedness of the distribution, compared with no label smoothing (nols), for German-to-English (see Appendix~\ref{appendix:label-smoothing} for Chinese-to-English). Label smoothing consistently increases entropy and decreases total probability mass across all values of $s$.}
     \label{fig:ls-peakedness}
\end{figure*}

\begin{figure*}
\centering
    \begin{subfigure}[b]{0.4\textwidth}
        \includegraphics[width=\textwidth,trim=15 15 10 10]{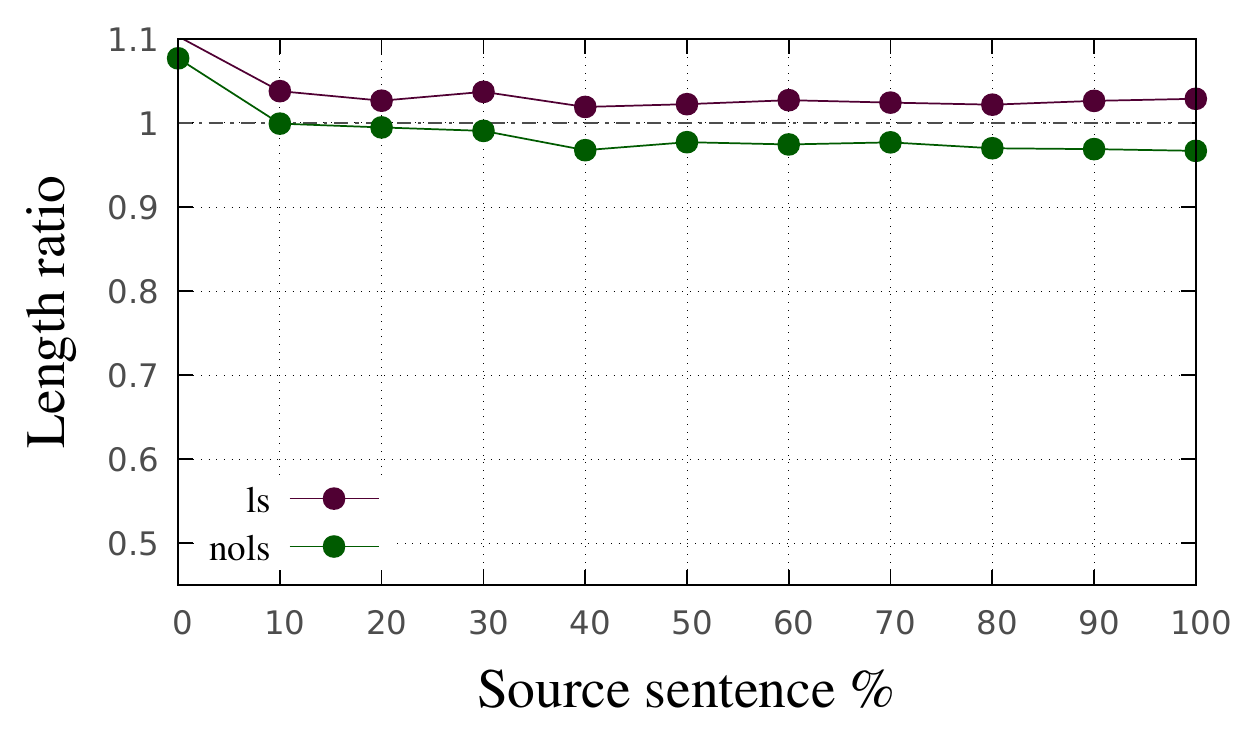}
        \caption{Length ratio, samples}
        \label{fig:ls-length-samples}
    \end{subfigure}%
    \hfill
    \begin{subfigure}[b]{0.4\textwidth}
        \includegraphics[width=\textwidth,trim=15 15 10 10]{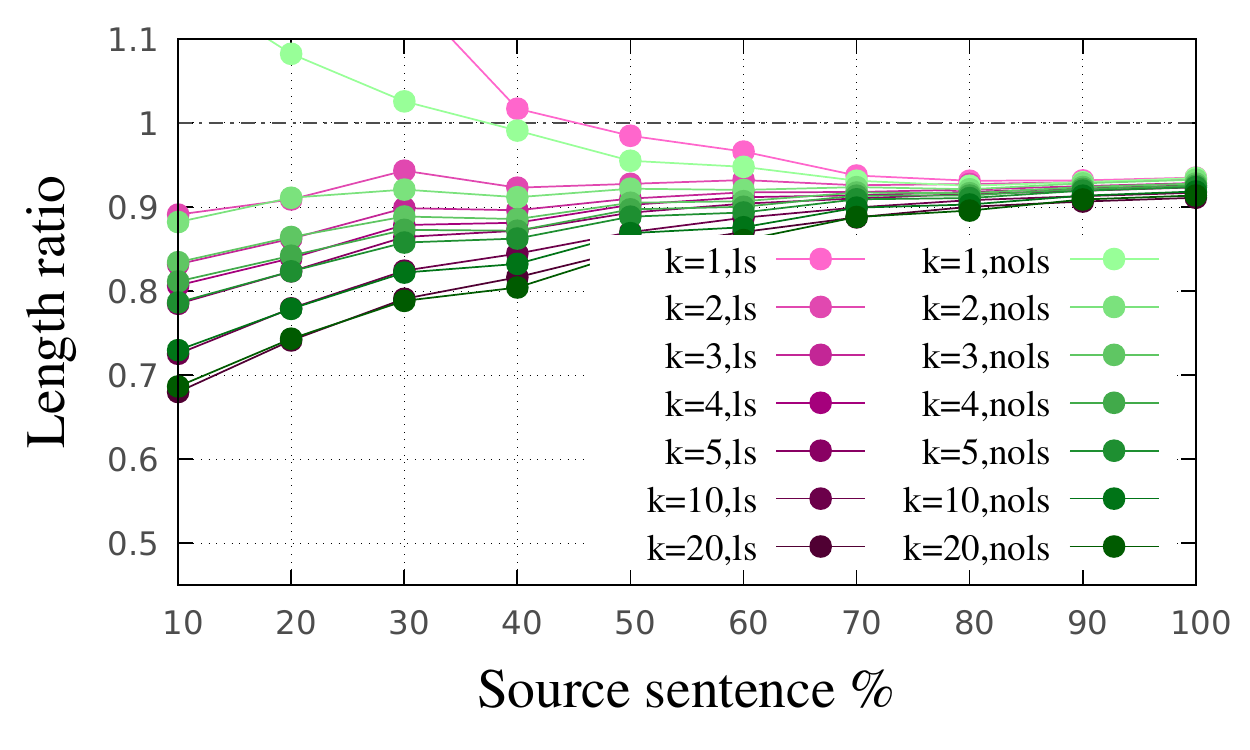}
        \caption{Length ratio, beam search}
        \label{fig:ls-length-search}
    \end{subfigure}
    \\
    \vspace{3ex}
    \begin{subfigure}[b]{0.4\textwidth}
        \includegraphics[width=\textwidth,trim=15 15 10 10]{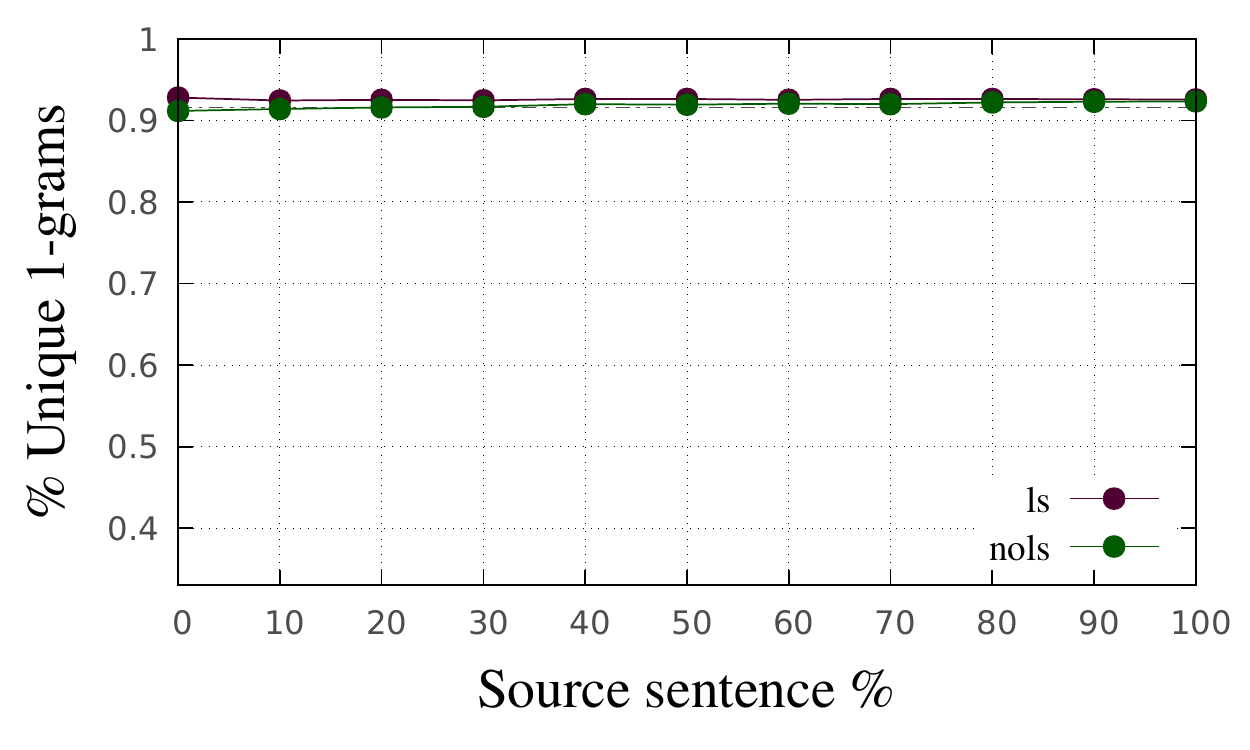}
        \caption{Repetition, samples}
        \label{fig:ls-repetition-samples}
    \end{subfigure}%
    \hfill
    \begin{subfigure}[b]{0.4\textwidth}
        \includegraphics[width=\textwidth,trim=15 15 10 10]{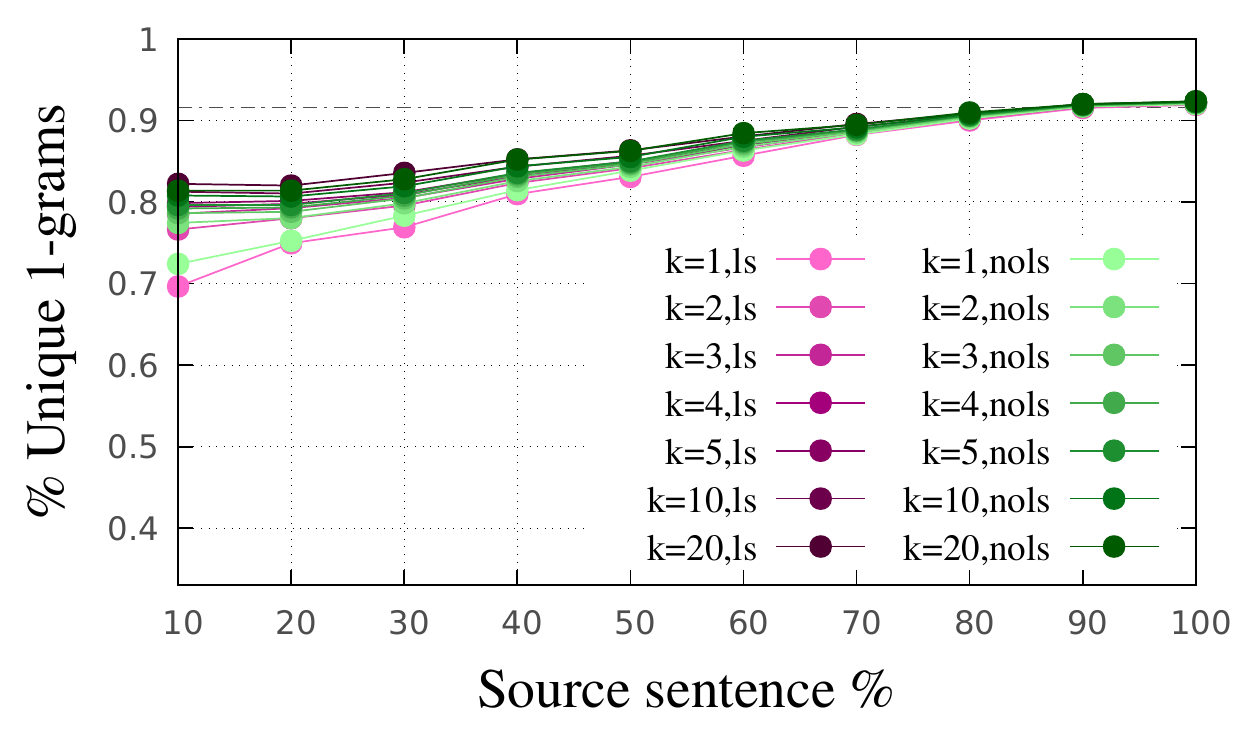}
        \caption{Repetition, beam search}
        \label{fig:ls-repetition-search}
    \end{subfigure}
    \caption{Length ratio of translations and percentage of unique 1-grams versus source sentence percentage ($s$), both with label smoothing (ls) and without (nols). Results for samples are computed based on 1000 samples for each test sentence; results for beam search vary across beam sizes ($k$). For samples, label smoothing increases the length ratio from slightly below the reference length to slightly above it; otherwise it has no discernible effect. (These results are for German-to-English; see Appendix~\ref{appendix:label-smoothing} for Chinese-to-English.)}
    \label{fig:ls-de}
\end{figure*}

\section{Label Smoothing and Degeneration}
\label{sec:label-smoothing}

We now begin to examine exactly what it is about task constrainedness which affects the amount of degeneration. One possible explanation is that, as we vary $s$, we vary the distribution's peakedness: the distribution becomes much less peaked as $s$ decreases (as shown in \S\ref{subsec:peakedness}). To examine whether differences in peakedness fully explain the level of degeneration, we contrast with a different method of adjusting peakedness: label smoothing.
Label smoothing \citep{szegedy+:2016} is an alternative to the standard cross-entropy loss function. Instead of comparing the next-word distribution against a one-hot vector, it compares against a mixture of a one-hot vector and the uniform distribution. It is commonly used in modern NMT systems, and has generally been found to be helpful, though the reasons why are still being investigated \citep{muller+:2019,lukasik+:2020,gao-etal-2020-towards}.

Label smoothing has the effect of smoothing the distribution over more output tokens at each timestep. This has a big effect on the peakedness, as shown in Fig~\ref{fig:ls-peakedness}. But, as we will show, it has almost no impact on either length bias or repetition. (All the graphs in this section show German-to-English only; the Chinese-to-English results are similar and can be found in Appendix~\ref{appendix:label-smoothing}.)

The biggest effect we see is in Figure~\ref{fig:ls-length-samples}, which shows how adding label smoothing impacts the length bias when sampling. Here, label smoothing changes the length bias from just below 1 to just above 1, giving sentences which are, on average, very slightly longer than the reference.

However, although label smoothing affects length bias for the overall distribution, we see essentially no effect on length bias when using beam search (Figure~\ref{fig:ls-length-search}).\footnote{We do note, however, that \citet{peters-martins-2021-smoothing} did find that label smoothing affected length bias in the mode of the distribution.} Similarly, Figures~\ref{fig:ls-repetition-samples} and \ref{fig:ls-repetition-search} show the effect of label smoothing on $1$-gram repetition, for both search and sampling; there is essentially no effect. (We found this to be true for other values of $n$ as well.)

From this, we can conclude that it is not merely the spread of the distribution which causes these degenerate behaviors to occur. There must be some other property of task constrainedness which is influencing them. We leave further investigation of what that property might be to future work.

\section{Conclusion}

We introduced a new experimental framework for directly controlling the level of task constrainedness, by truncating sentences on the source side of an MT system. Using this experimental framework, we analyzed how task constrainedness affected degenerate behaviors.

For less constrained tasks, we observe three failure modes: beam search decoding that is too short, greedy decoding that is too long and repetitive, and random samples that are disfluent. We note that the same three failure modes are also displayed by a simple unigram language model: since every sentence contains \eos{}, the highest-probability output must be empty (just \eos{} with no real words); since $P(\eos) < P(\text{the})$, a greedy search will choose \emph{the} over and over; and random samples from a unigram distribution are of course disfluent. So the simplest explanation may be that the neural models used here are still insufficiently sensitive to context.

For more constrained tasks, these effects are much milder. The presence of the source sentence seems to be sufficient to all but eliminate repetition and noticeably improve fluency.
Although some work on RNN models for NMT focused on adding coverage models to reduce skipping and repeating of source words \citep{tu-etal-2016-modeling,mi-etal-2016-coverage,li-etal-2018-simple}, Transformers seem to suffer from these problems far less.
As Transformers were originally designed for the $s=100$ case, one direction for future research may be to investigate modifications of the Transformer that are better-suited to less constrained tasks.

\section*{Acknowledgements}

We thank Toan Nguyen for his help with the \verb|transformers-without-tears| library, and Ari Holtzman for useful discussion.

This material is based upon work supported by the National Science Foundation under Grant No.~CCF-2019291. Any opinions, findings, and conclusions or recommendations expressed in this material are those of the authors and do not necessarily reflect the views of the National Science Foundation.

\bibliography{anthology,custom}
\bibliographystyle{acl_natbib}

\clearpage
\appendix

\section{Additional results on repetition}
\label{appendix:repetition}

\begin{figure*}
\centering
    \begin{subfigure}[b]{0.45\textwidth}
        \includegraphics[width=\textwidth,trim=15 15 10 10]{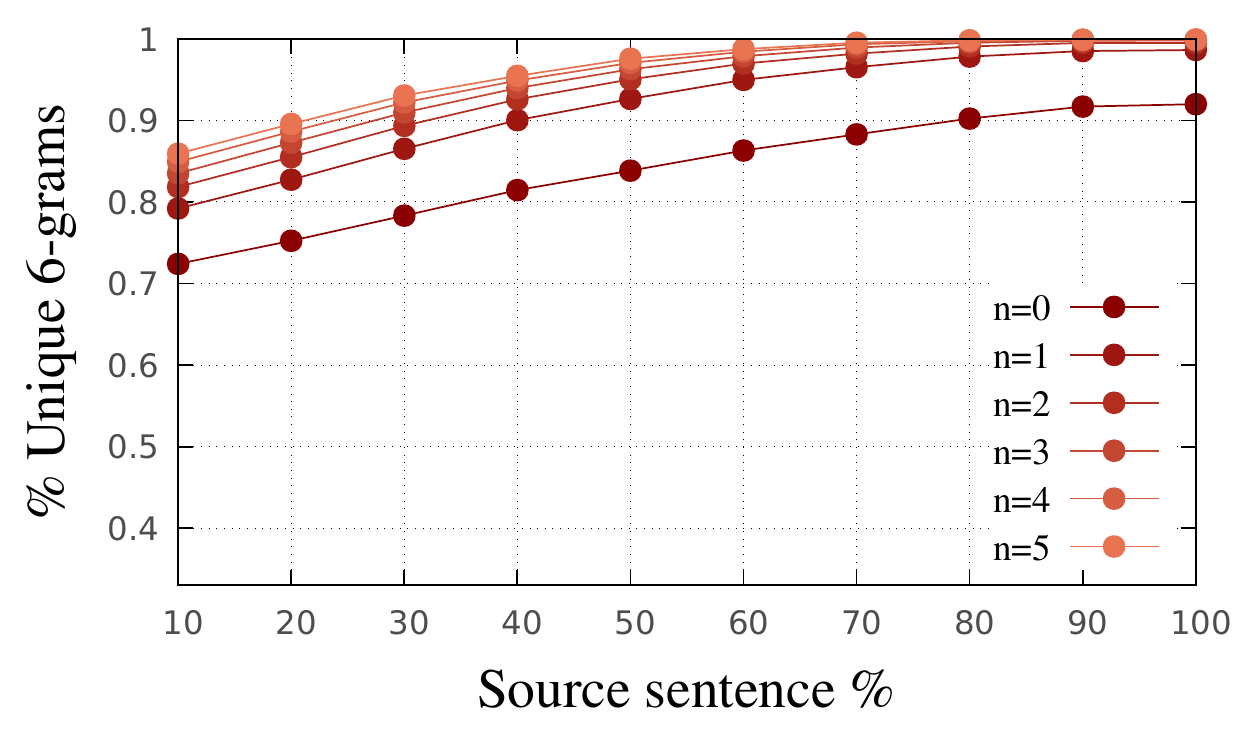}
        \caption{German--English}
    \end{subfigure}%
    \hfill
    \begin{subfigure}[b]{0.45\textwidth}
        \includegraphics[width=\textwidth,trim=15 15 10 10]{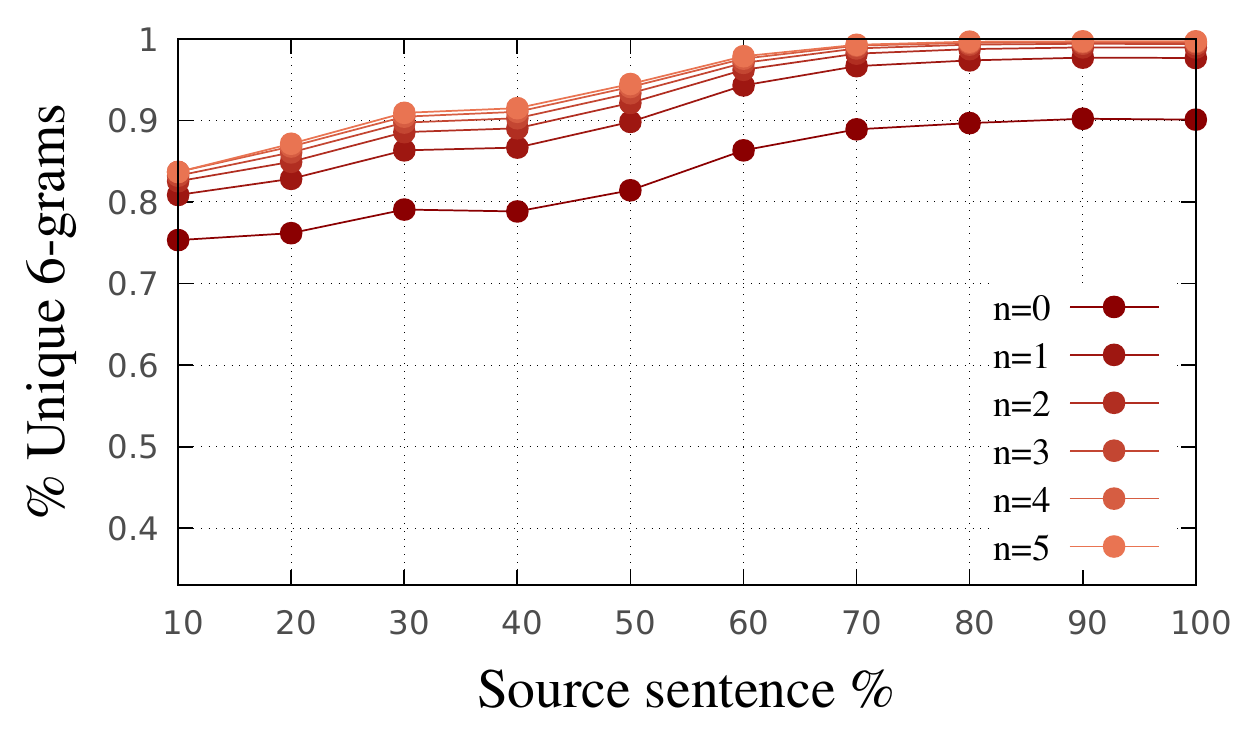}
        \caption{Chinese--English}
    \end{subfigure}%
    \caption{Amount of repetition, measured as the percentage of $n$-grams in the sentence which are unique, versus source sentence percentage ($s$). This is mostly the same information shown in Figures \ref{fig:repetition-de} and \ref{fig:repetition-zh}, but viewed in a different way: here, we look at just one beam size ($k=1$, for which the repetition was most pronounced), and compare multiple $n$. All values of $n$ show a similar pattern, with considerable repetition observed even for 6-grams for low $s$.}
\label{fig:repetition-n}
\end{figure*}

\begin{figure*}
\centering
    \begin{subfigure}[b]{0.45\textwidth}
        \includegraphics[width=\textwidth,trim=15 15 10 10]{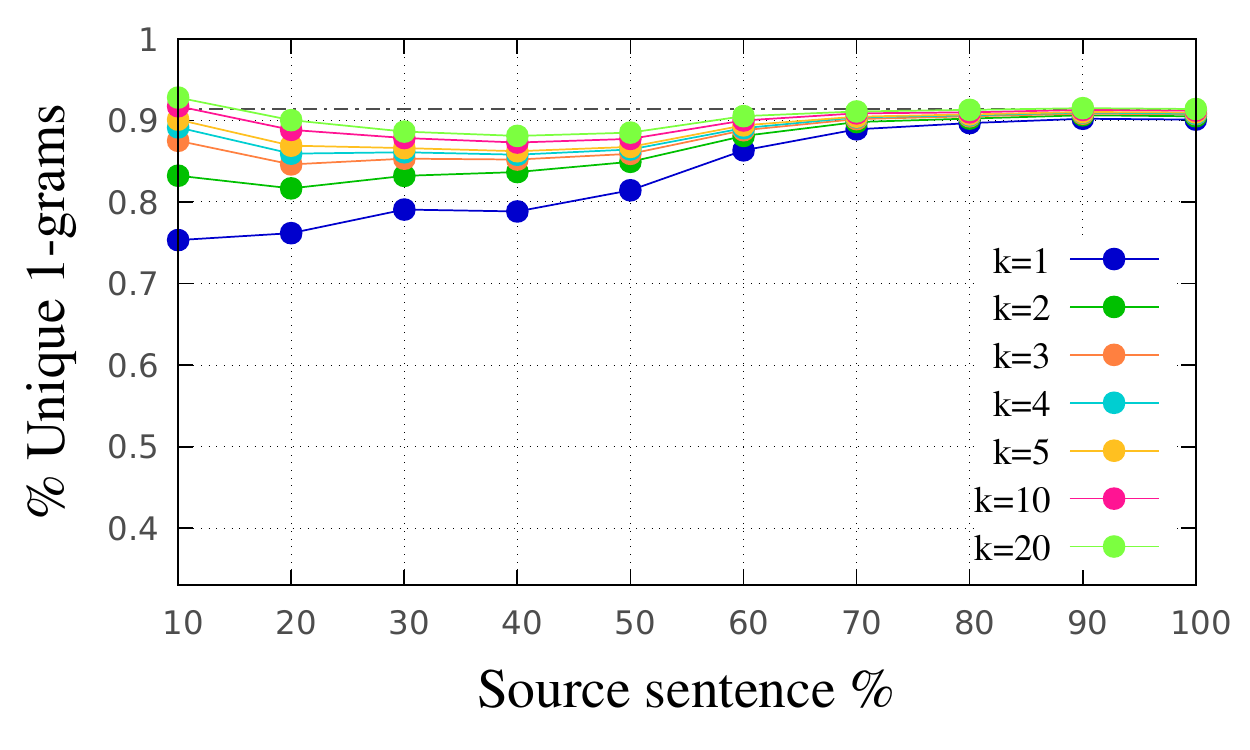}
    \end{subfigure}%
    \hfill
    \begin{subfigure}[b]{0.45\textwidth}
        \includegraphics[width=\textwidth,trim=15 15 10 10]{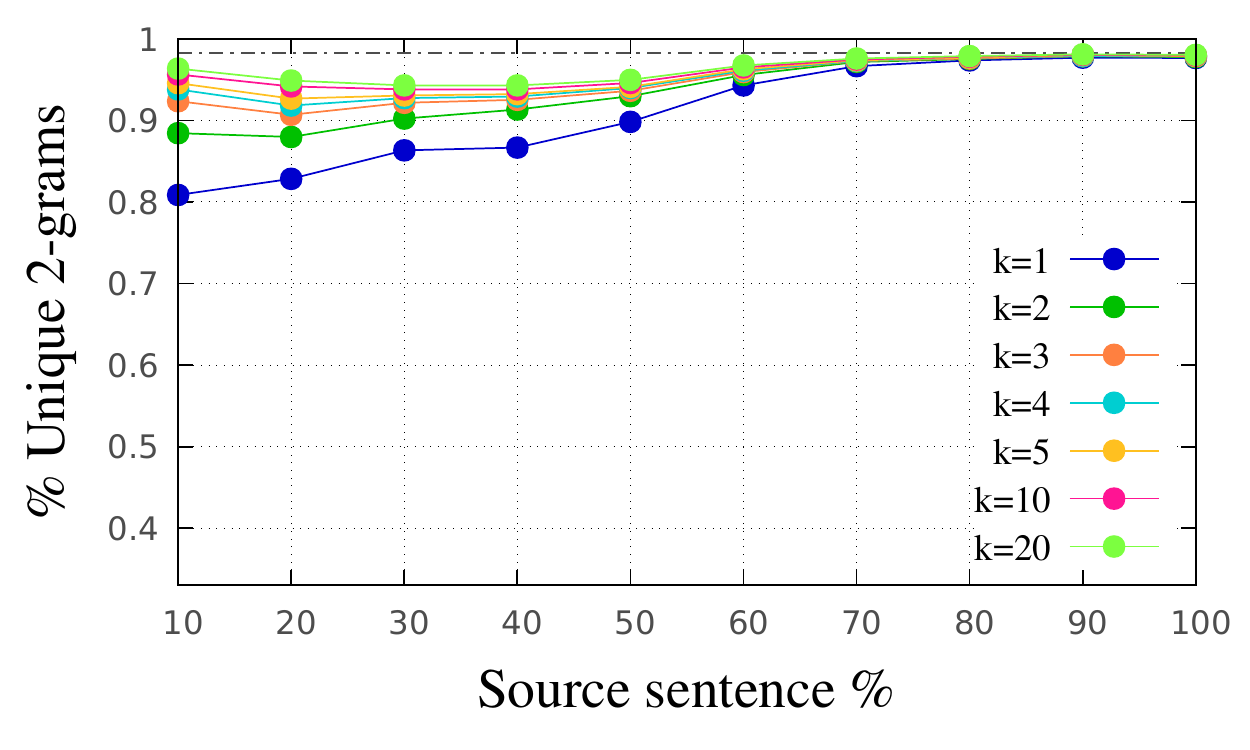}
    \end{subfigure}
    \\ \vspace{-0.73cm}
    \vspace{3ex}
    \begin{subfigure}[b]{0.45\textwidth}
        \begin{mdframed}[backgroundcolor=white,linewidth=0pt,innertopmargin=0pt,innerbottommargin=0pt,innerleftmargin=0pt,innerrightmargin=0pt]
        \includegraphics[width=\textwidth,trim=15 15 10 10]{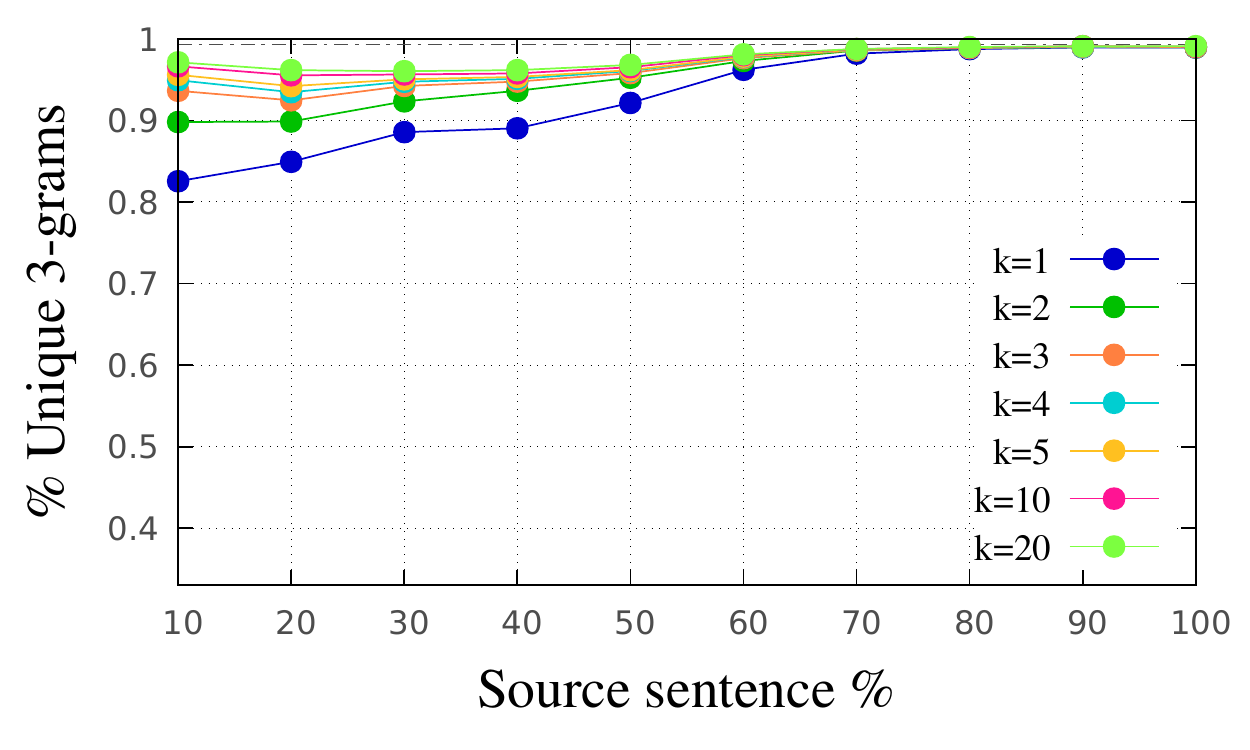}
        \end{mdframed}
    \end{subfigure}%
    \hfill
    \begin{subfigure}[b]{0.45\textwidth}
        \begin{mdframed}[backgroundcolor=white,linewidth=0pt,innertopmargin=0pt,innerbottommargin=0pt,innerleftmargin=0pt,innerrightmargin=0pt]
        \includegraphics[width=\textwidth,trim=15 15 10 10]{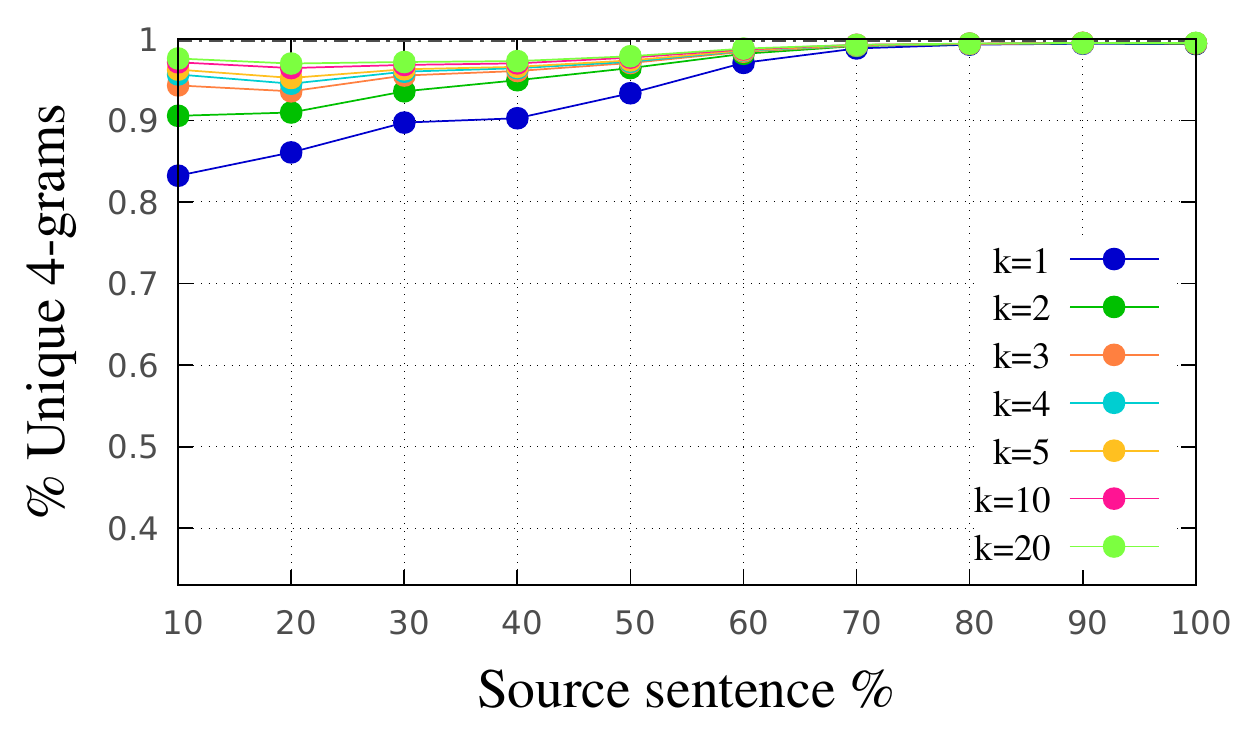}
        \end{mdframed}
    \end{subfigure}
    \caption{Amount of repetition versus source sentence percentage ($s$), for various beam sizes ($k$). (Graphs show Chinese-to-English results only.) Repetition is measured as the percentage of unique $n$-grams in a sentence; the graphs show this for different values of $n$. The repetition rate of the reference is plotted as a dashed grey line. As in German-to-English, the percent of unique $n$-grams drops as $s$ decreases across all values of $n$, while repetition actually becomes less of a problem for higher values of $k$.}
    \label{fig:repetition-zh}
\end{figure*}

Here we show some additional outputs from our systems. Figure~\ref{fig:repetition-zh} graphs the amount of repetition for the Chinese-to-English systems; we see similar results to the German-to-English systems, but with an even more pronounced decrease in repetition for higher beam sizes $k$.

Figure~\ref{fig:repetition-n} displays the same results, but graphed in terms of $n$. (We look at beam size $k=1$ since repetition is most pronounced in that case.) This graph shows a surprising consistency across $n$; although the effect is most pronounced for 1-gram repetition, we still see quite a bit of degenerate repetition even up to $6$-grams, suggesting that the phrases which are being repeated are quite long.

\section{Additional outputs from our model}
\label{appendix:examples}

As a supplement to Table~\ref{tab:ex1}, we present some additional outputs from our system, which show similar trends.

\begin{table*} \centering \scriptsize
\begin{tabular}{@{}rl@{}}
\toprule
$s$ (\%) & Output found using beam search, $k=4$ \\
\midrule
0 & And I said, "Well, I'm going to show you a little bit." \\
10 & A few years ago, I was in the hospital, and I was in the hospital. \\
20 & A few years ago, when I was a kid, I was a kid. \\
30 & A few years ago, here at TED, I'm going to tell you a little bit about this. \\
40 & A couple of years ago, at TED, I'm going to tell you a little bit about this. \\
50 & A couple of years ago, at TED, Peter Peter asked me, "What are you doing?" \\
60 & A couple of years ago, here at TED, Peter Skillman introduced a book called "The Sun." \\
70 & A couple of years ago, here at TED, Peter Skillman introduced a design competition called "The House." \\
80 & A few years ago, here at TED, Peter Skillman introduced a design competition called "The Government." \\
90 & A few years ago, here at TED, Peter Skillman made a design competition called "The Marshmallow Child." \\
100 & A few years ago, here at TED, Peter Skillman introduced a design competition called "The Marshmallow Child." \\
\midrule
ref & Several years ago here at TED, Peter Skillman  introduced a design challenge  called the marshmallow challenge. \\
\bottomrule
\end{tabular}
\caption{Beam search ($k=4$) outputs for a sentence in the test dataset, shown across all values of $s$.}
\label{tab:ex2}
\end{table*}

\begin{table*} \centering \scriptsize
\begin{tabular}{@{}rl@{}}
\toprule
$s$ (\%) & Output found using beam search, $k=4$ \\
\midrule
0 & And I said, "Well, I'm going to show you a little bit." \\
10 & A child, a child, a child, a child, a child, a child, a child, a child. \\
20 & A child who is living in the world today is a child, a child, a child, a child, a child, a child, a child. \\
30 & A child who's born in New Delhi today will be born in a new world, a new world, a new world, a new world. \\
40 & A child born today in New Delhi can expect to be a child who has been born in the United States. \\
50 & A child who's born in New Delhi today can expect to be as long as they're born, and that's where they are. \\
60 & A kid who can be born in New Delhi today would expect to live as long as they were, and that's what they were doing. \\
70 & A child born in New Delhi today will expect to live as long as the richest child in the world. \\
80 & A child born in New Delhi today will expect to live as long as the richest man on the planet. \\
90 & A child born today in New Delhi can expect to live as long as the richest man in the world, 100 years ago. \\
100 & A child born today in New Delhi can expect to live as long as the richest man in the world 100 years ago. \\
\midrule
ref & A kid born in New Delhi today can expect to live as long as the richest man in the world did 100 years ago. \\
\bottomrule
\end{tabular}
\caption{Beam search ($k=4$) outputs for a sentence in the test dataset, shown across all values of $s$.}
\label{tab:ex3}
\end{table*}

\begin{table*} \centering \scriptsize
\begin{tabular}{@{}rl@{}}
\toprule
$s$ (\%) & Output found using beam search, $k=4$ \\
\midrule
0 & And I said, "Well, I'm going to show you a little bit." \\
10 & Today, I'm going to show you a couple of examples of how this works. \\
20 & Today, I don't know how many of you have heard of this, but I think it's a very important thing. \\
30 & Today, I don't know, maybe it's going to go up today, but I'm going to tell you a little bit about what I'm going to talk about today. \\
40 & Now, I don't know, maybe someone in this room came to me and said, "You know, I'm going to go to school." \\
50 & Today, I don't know, maybe someone in Lagos came to a place where I was born, and I didn't know what to do. \\
60 & Today, I don't know, maybe somebody in Lagos came into an airplane this morning, and it's going to be like, "Oh, I'm going to die." \\
70 & Now, I don't know, maybe someone in Lagos this morning went into an airplane, and it's going to go to Los Angeles, and it's going to be a big deal. \\
80 & Today, I don't know, maybe someone in Lagos came to an airplane this morning, and it's just going to Los Angeles, in the middle of the Central Sea. \\
90 & Today, I don't know, maybe someone in Lagos is going to go to an airplane this morning, and he's going to go to Los Angeles, right now he's gone. \\
100 & Today, I don't know, maybe someone in Lagos climbed into an airplane this morning, and it's just going to Los Angeles right now, it's over Ohio. \\
\midrule
ref & Today, I don't know, maybe a guy got on a plane in Lagos this morning, and he's flying to LAX, right now he's over Ohio. \\
\bottomrule
\end{tabular}
\caption{Beam search ($k=4$) outputs for a sentence in the test dataset, shown across all values of $s$.}
\label{tab:ex4}
\end{table*}

\begin{table*} \centering \scriptsize
\begin{tabular}{@{}rl@{}}
\toprule
$s$ (\%) & Output found using beam search, $k=4$ \\
\midrule
0 & And I said, "Well, I'm going to show you a little bit." \\
10 & If you look at it, you can see that it's a little bit different. \\
20 & If you're 10 teams, you're going to have to be able to do that. \\
30 & If you have 10 teams, you have 10 teams, and you have 10 teams, and you have them. \\
40 & If you have 10 teams, typically, you have 10 teams, and you have 10 teams. \\
50 & If you have 10 teams that are typically predicting, you're not going to be able to do that. \\
60 & If you have 10 teams that are typically predicted, you get 10 teams, and you get 10 teams. \\
70 & If you have 10 teams that typically go, you get about six teams per second. \\
80 & If you have 10 teams that typically go ahead, you get about six, the two teams. \\
90 & If you have 10 teams that are typical, you get about six, the stable structures. \\
100 & If you have 10 teams that go typically, you get about six that have stable structures. \\
\midrule
ref & If you have 10 teams that typically perform, you'll get maybe six or so that have standing structures. \\
\bottomrule
\end{tabular}
\caption{Beam search ($k=4$) outputs for a sentence in the test dataset, shown across all values of $s$.}
\label{tab:ex5}
\end{table*}

\section{Additional results on label smoothing}
\label{appendix:label-smoothing}

Here we present additional results on label smoothing, for the Chinese-to-English language pair. These results are quite similar to the ones observed for German-to-English. Again, we see a substantial difference in the peakedness of the distribution. And again, we notice a slight change in length ratio for the samples, but otherwise, we observe essentially no effect of label smoothing on degenerate behavior.

\begin{figure*}
     \centering
     \begin{subfigure}[b]{0.31\textwidth}
         \includegraphics[width=\textwidth,trim=15 15 15 15]{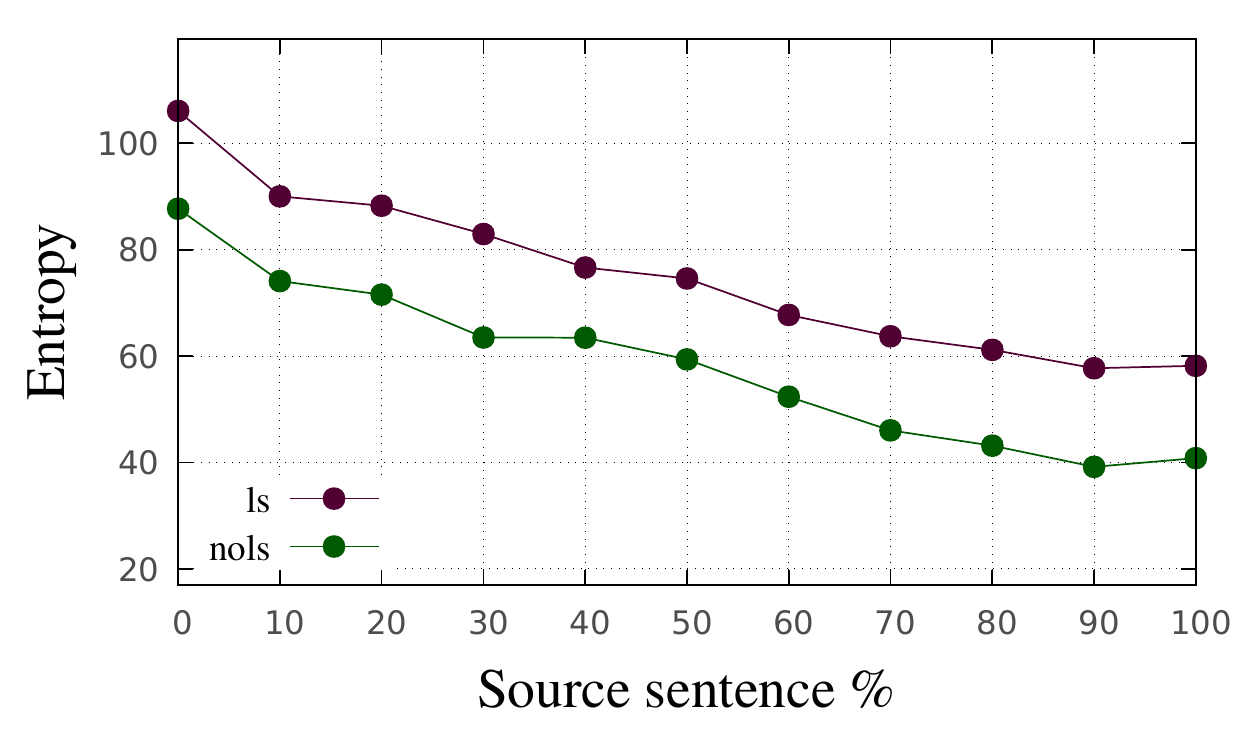}
         \caption{Per-sentence entropy (nats), zh-en}
     \end{subfigure}%
     \hfill
     \begin{subfigure}[b]{0.31\textwidth}
         \includegraphics[width=\textwidth,trim=15 15 15 15]{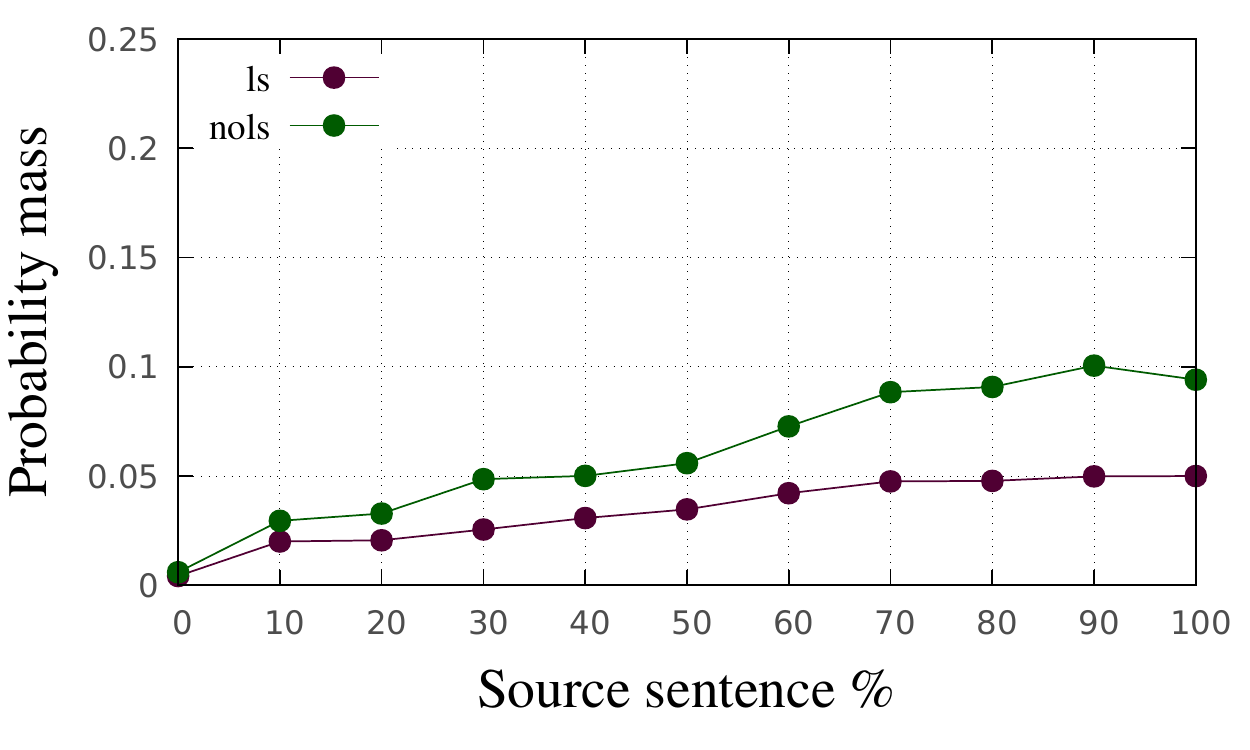}
         \caption{Total probability mass, zh-en}
     \end{subfigure}%
     \hfill
     \begin{subfigure}[b]{0.31\textwidth}
         \includegraphics[width=\textwidth,trim=15 15 15 15]{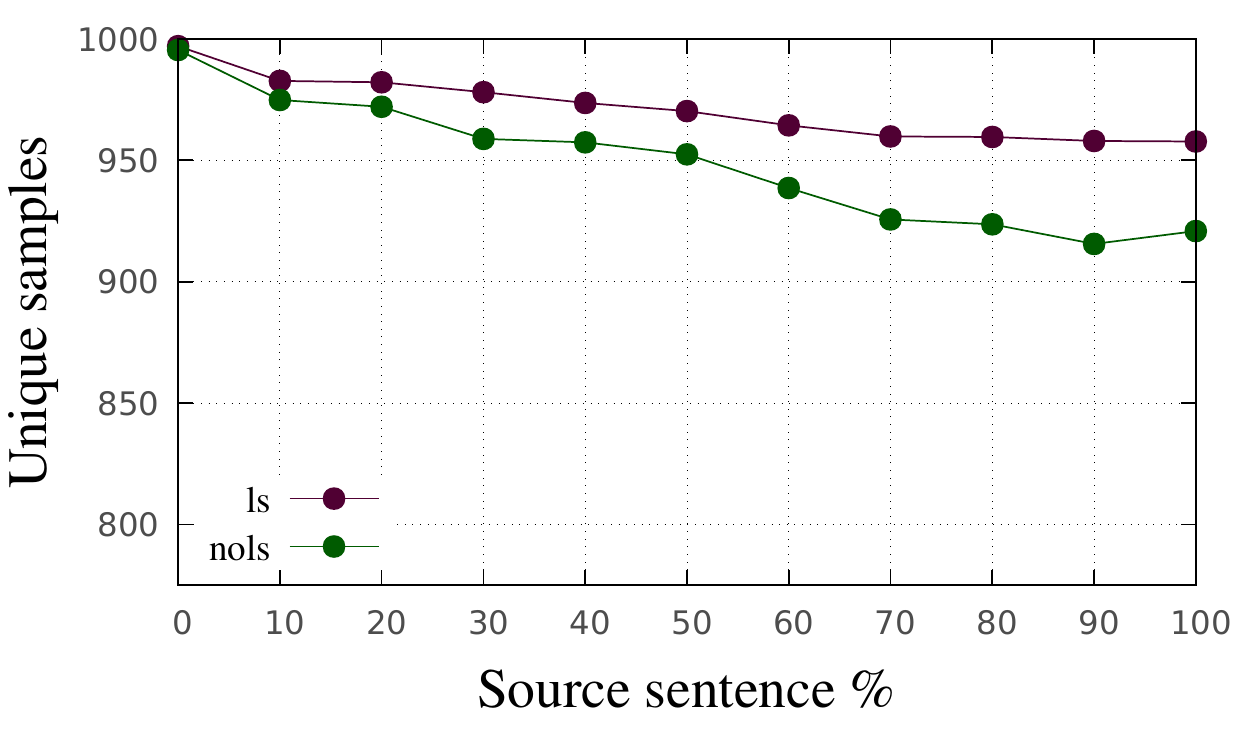}
         \caption{Number of unique samples, zh-en}
     \end{subfigure}%
     \caption{Effect of label smoothing (ls) on the peakedness of the distribution, compared with no label smoothing (nols), for Chinese-to-English. As with German-to-English, label smoothing consistently increases entropy and decreases total probability mass across all values of $s$.}
     \label{fig:ls-peakedness-zh}
\end{figure*}

\begin{figure*}
\centering
    \begin{subfigure}[b]{0.4\textwidth}
        \includegraphics[width=\textwidth,trim=15 15 10 10]{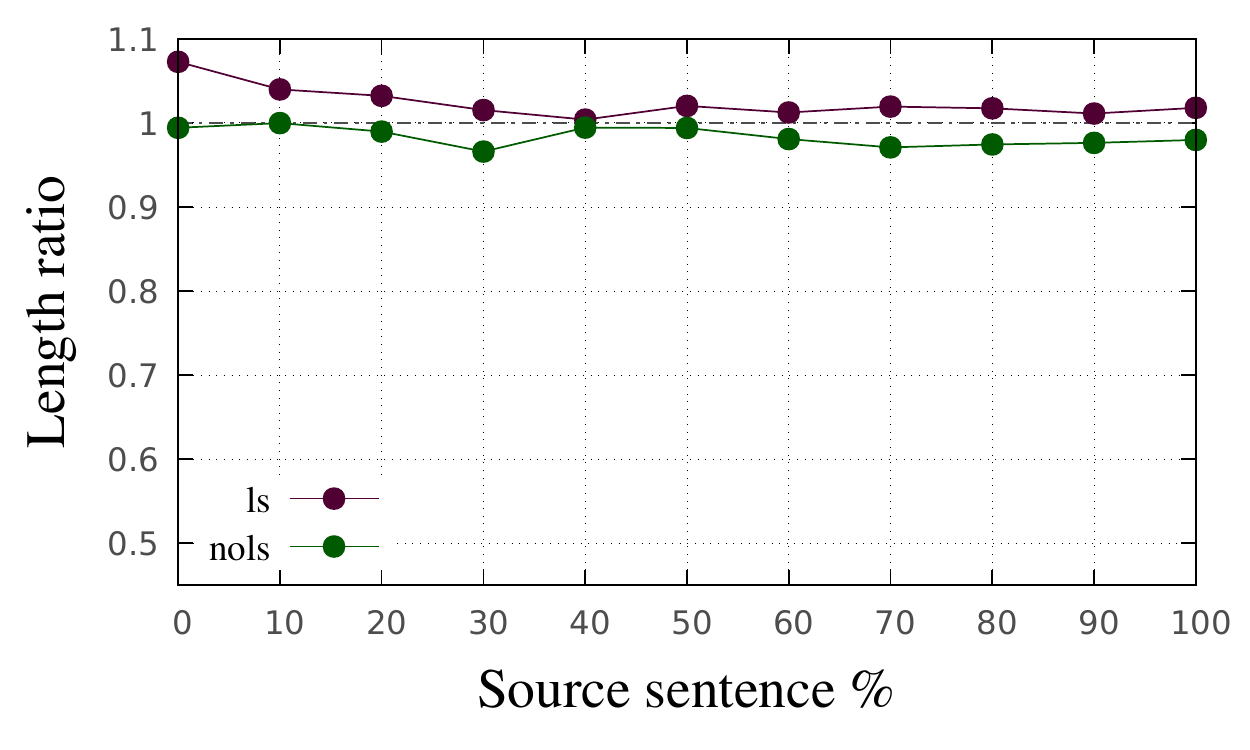}
        \caption{Length ratio, samples}
    \end{subfigure}%
    \hfill
    \begin{subfigure}[b]{0.4\textwidth}
        \includegraphics[width=\textwidth,trim=15 15 10 10]{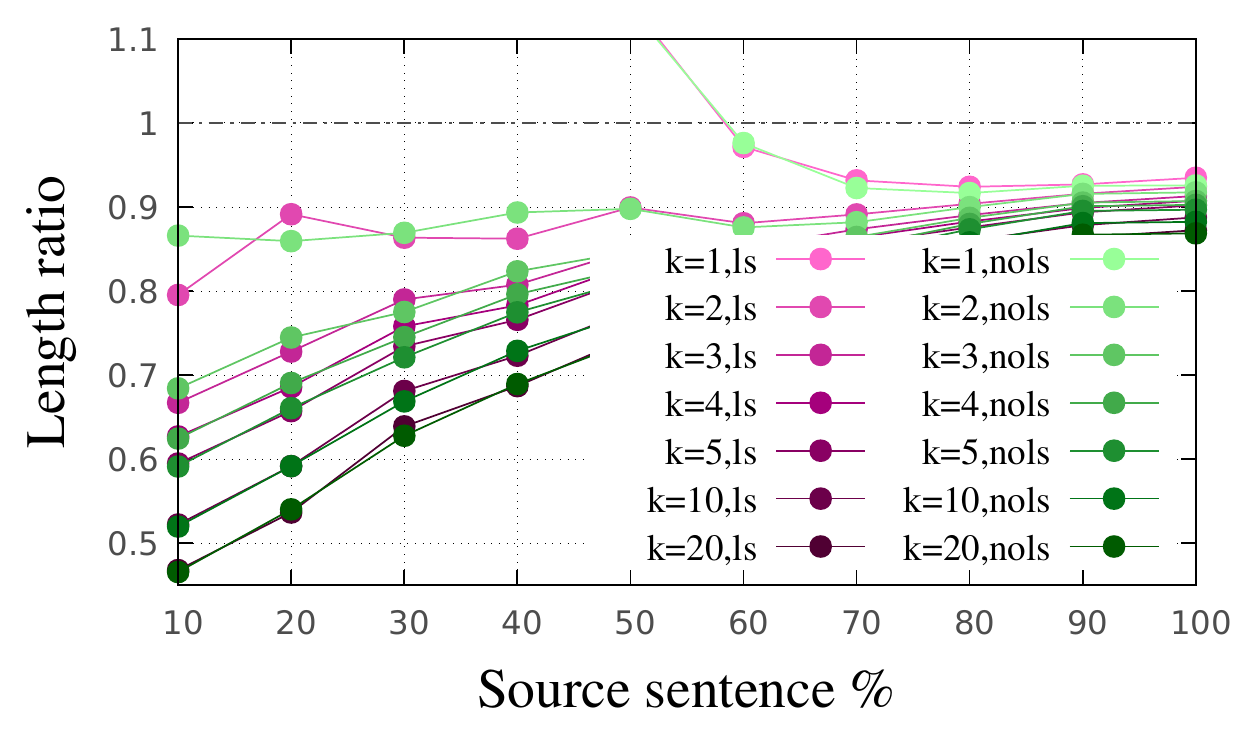}
        \caption{Length ratio, beam search}
    \end{subfigure}
    \\
    \vspace{3ex}
    \begin{subfigure}[b]{0.4\textwidth}
        \includegraphics[width=\textwidth,trim=15 15 10 10]{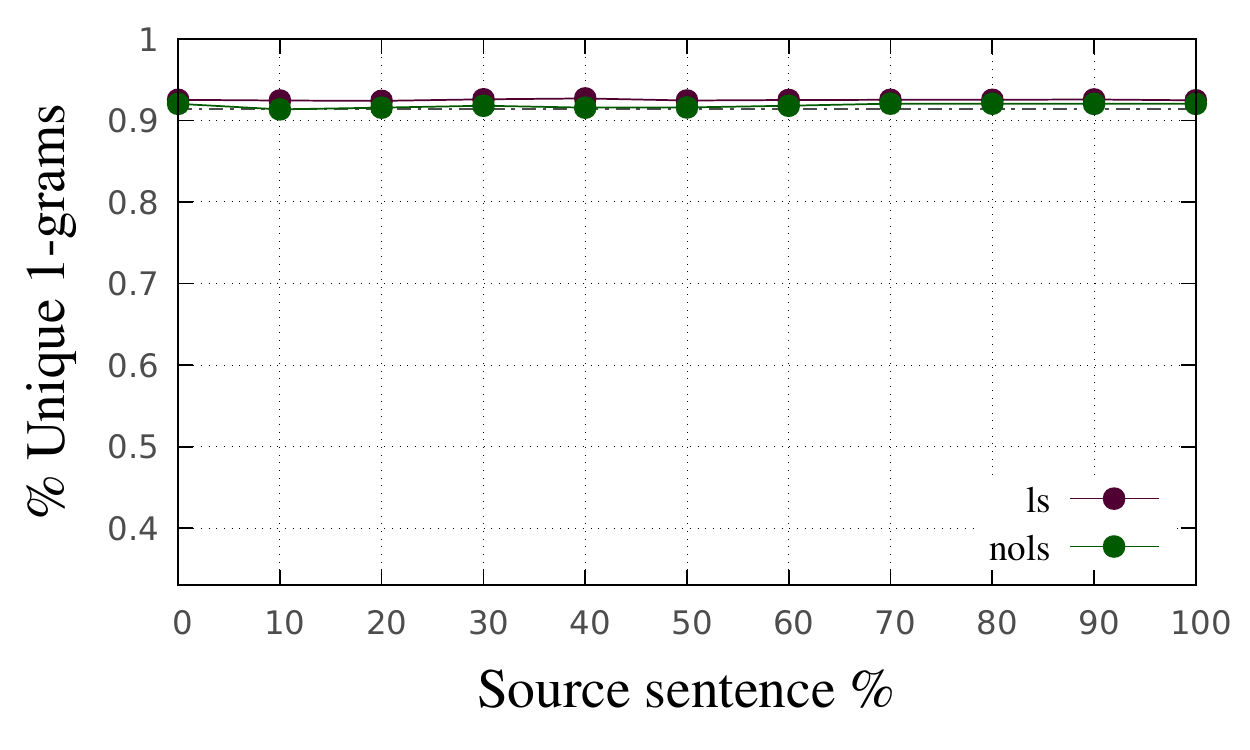}
        \caption{Repetition, samples}
    \end{subfigure}%
    \hfill
    \begin{subfigure}[b]{0.4\textwidth}
        \includegraphics[width=\textwidth,trim=15 15 10 10]{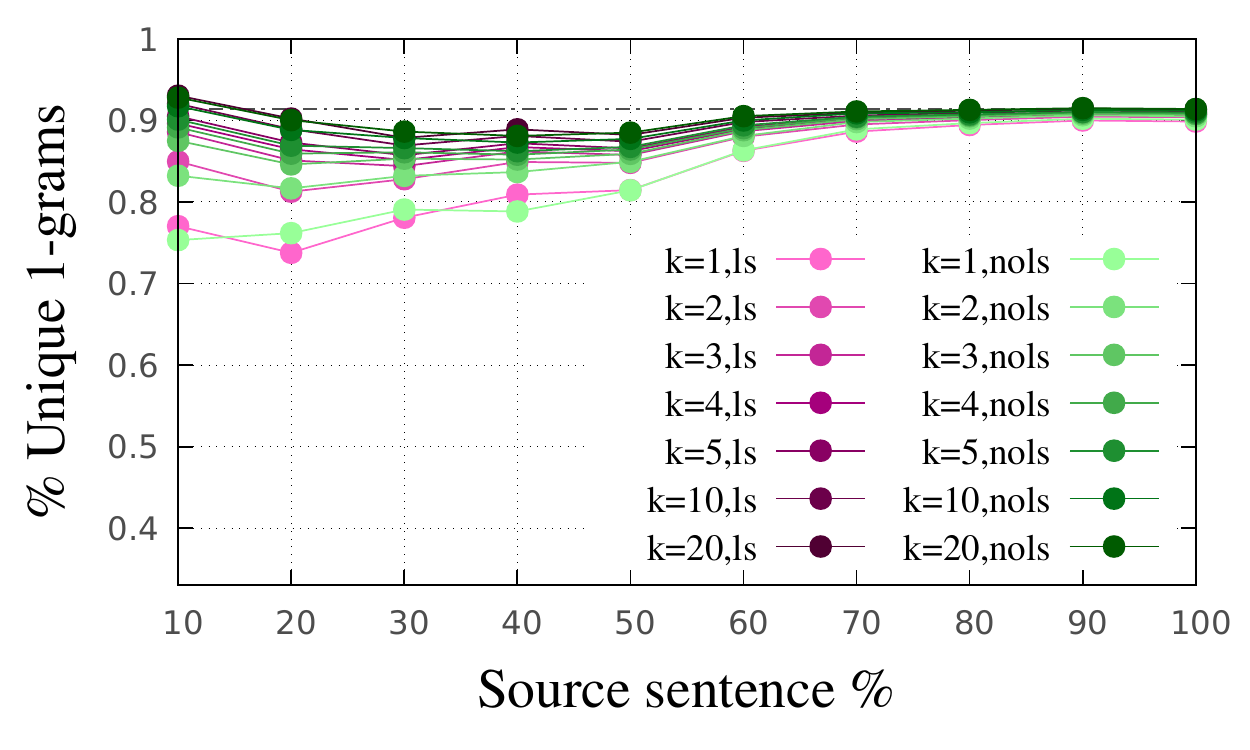}
        \caption{Repetition, beam search}
    \end{subfigure}
    \caption{Length ratio of translations and percentage of unique 1-grams versus source sentence percentage ($s$), both with label smoothing (ls) and without label smoothing (nols). Results for samples are computed based on 1000 samples for each test sentence; results for beam search vary across beam sizes ($k$). As with the German-to-English results, we find that, for samples, label smoothing increases the length ratio from slightly below the reference length to slightly above it; otherwise it has no discernable effect.}
    \label{fig:ls-zh}
\end{figure*}

\end{document}